\documentclass[preprint,12pt]{elsarticle}

\usepackage{graphicx}
\usepackage{amssymb}

\usepackage{lineno}
\usepackage{url}
\usepackage{caption}
\DeclareCaptionLabelFormat{cont}{#1~#2\alph{ContinuedFloat}}
\usepackage[linesnumbered,ruled,vlined]{algorithm2e}
\usepackage{makecell}
\usepackage{graphicx}
\usepackage{rotating}
\usepackage{pdflscape}

\journal{Elsevier}

\begin{document}
    \nolinenumbers
	
	\begin{frontmatter}
		
		
		\title{Deep Merge: Deep-learning-based Region Merging for Remote Sensing Image Segmentation}
		

		
		\author[a,b]{Xianwei Lv}
		\author[c]{Claudio Persello}
		\author[d]{Wangbin Li\corref{cor1}}
		\author[e]{Xiao Huang}
		\author[f]{Dongping Ming}
		\author[c]{Alfred Stein}
		
		\address[a]{School of Computer and Communication Engineering, Northeastern University at Qinhuangdao,  Qinhuangdao 066004, China}
		\address[b]{Hebei Key Laboratory of Marine Perception Network and Data Processing, Northeastern University at Qinhuangdao 066004, China.}
		\address[c]{Dept. of Earth Observation Science, Faculty ITC, University of Twente, 7500AE Enschede, the Netherlands}
		\address[d]{State Key Laboratory of Information Engineering in Surveying, Mapping and Remote Sensing, Wuhan University, Wuhan 430079, China}
		
		\address[e]{The Department of Geosciences, University of Arkansas, Fayetteville, AR, USA}
		\address[f]{School of Information Engineering, China University of Geosciences (Beijing), Beijing 100083, China}
		\cortext [cor1]{Corresponding author. E-mail address: lwb1996@whu.edu.cn (Wangbin Li)}
		
		\begin{abstract}
			Image segmentation aims to partition an image according to the objects in the scene and is a fundamental step in analysing very high spatial-resolution (VHR) remote sensing imagery. Current methods struggle to effectively consider land objects with diverse shapes and sizes. Additionally, the determination of segmentation scale parameters frequently adheres to a static and empirical doctrine, posing limitations on the segmentation of large-scale remote sensing images and yielding algorithms with limited interpretability. To address the above challenges, we propose a deep-learning-based region merging method dubbed DeepMerge to handle the segmentation of complete objects in large VHR images by integrating deep learning and region adjacency graph (RAG). This is the first method to use deep learning to learn the similarity and merge similar adjacent super-pixels in RAG. We propose a modified binary tree sampling method to generate shift-scale data, serving as inputs for transformer-based deep learning networks, a shift-scale attention with 3-Dimension relative position embedding to learn features across scales, and an embedding to fuse learned features with hand-crafted features. DeepMerge can achieve high segmentation accuracy in a supervised manner from large-scale remotely sensed images and provides an interpretable optimal scale parameter, which is validated using a remote sensing image of 0.55 m resolution covering an area of 5,660 km$^{2}$. The experimental results show that DeepMerge achieves the highest \emph{F} value (0.9550) and the lowest total error \emph{TE} (0.0895), correctly segmenting objects of different sizes and outperforming all competing segmentation methods.
		\end{abstract}
		
		
	\end{frontmatter}
	
	
	\section{Introduction}
	\label{S:1}
	
	Acquiring very high-spatial-resolution (VHR) remote sensing images over large areas has become easier than ever due to the advancement in satellite remote sensing technology. VHR images provide rich spatial details for characterizing objects on the ground. For this reason, they are widely applied in land-cover and land-use classification \cite{lv2021improved}, urban functional zone understanding \cite{zhou2020so}, urban management \cite{mundia2005analysis}, building roof modelling \cite{zhao2021building}, large-scale terrain classification \cite{na2021object}, and object extraction and monitoring \cite{chen2021urban,zhao2022extracting}. Objects in the land-cover classes tend to show high intra-class discrepancy and inter-class consistency, leading to challenges in image interpretation. Geographic Object-based Image Analysis (GeOBIA) has been proven to be an effective approach to address this issue by transitioning image interpretation from pixel-level to object-level using image segmentation, which clusters pixels into meaningful complete objects \cite{blaschke2010object}. Segmentation is different from semantic segmentation (or pixel-wise classification), which aims to classify each pixel in an image into specific categories and instance segmentation, whose goal is to identify and distinguish individual object instances within an image, providing pixel-level masks for each object. Image segmentation plays a crucial role in GeOBIA as well as other image analysis workflows. Objects are generally clustered with adjacent pixels with similar characteristics, such as spectral reflectance. Many efforts have been made to design precise segmentation methods and interpretable scale parameters for these methods. However, these methods fail to select a 'good' scale parameter to generate desirable segmentation results, as they tend to exhibit segmentation bias, including over-segmentation and under-segmentation errors, especially when applied to large-area VHR images. It is thus paramount to design more accurate and efficient segmentation algorithms with interpretable scale parameters that can be applied to a wide range of applications.\par
	
\begin{figure}[h]
		\centering\includegraphics[width=1.0\linewidth]{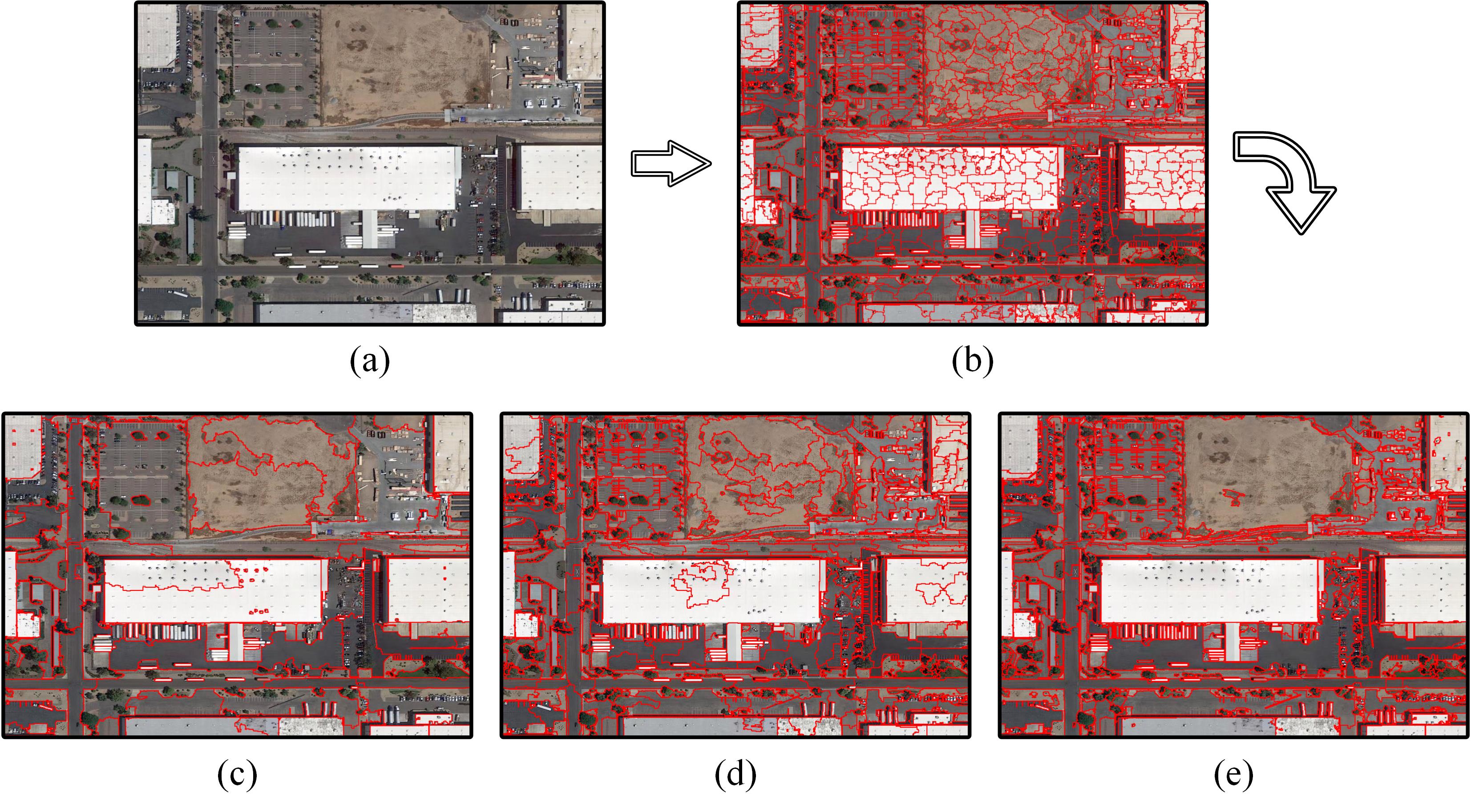}
		\caption{Comparison of different region-merging based segmentation methods in region-merging. (a) Original remote sensing image. (b) Initial over-segmented super-pixels. (c) Region-merging results of SOTA method BCMS \cite{zhang2013boundary}. (d) Region-merging results of SOTA method Local-SA \cite{yang2017region}. (e) Region-merging results of the proposed method DeepMerge.}
		\label{intro-goal}
	\end{figure}

	To satisfy the high segmentation quality demanded by diverse applications, region-merging-based segmentation methods were proposed, which will not segment image directly but initialize over-segmented super-pixels as primitives and further merge them into complete objects. Following spilt-and-merge, many efforts have been made to design various merging criteria to merge super-pixels into meaningful complete objects by learning similarities between neighbouring super-pixel pairs \cite{zhang2014fast}. The method involves three key steps: 1) initializing over-segmented super-pixels, 2) building a region adjacency graph (RAG) model \cite{beaulieu1989hierarchy,haris1998hybrid}, and 3) merging the most-similar super-pixel pairs according to the least weight edge extracted from the RAG. Fig.\ref{intro-goal} illustrates the segmentation results of the proposed method against state-of-the-art (SOTA) region-merging-based segmentation methods based on RAG.\par
	
	Region-merging-based methods via RAG can be implemented in an unsupervised or supervised manner. For unsupervised methods, users need to select a proper scale parameter (i.e., the threshold or the stopping rule) designed to stop merging by measuring discrepancy (or similarity) between adjacent super-pixel pairs. However, the scale parameter selection overly requires user experience and often introduces serious uncertainties in the segmentation results. Optimal scale parameter selection methods failed to adapt to various merging criteria, remote sensing sensors, and applications. Their segmentation results are far from satisfactory. Supervised segmentations are designed to measure the discrepancy or similarity among super-pixels via combined features between super-pixels and reference objects. However, supervised segmentations face challenges when applied to large VHR images. Most of the existing supervised methods are designed for a single scene, failing to generalize to other scenes due to the discrepancy in objects’ characteristics. \par
	
\begin{figure}[h]
		\centering\includegraphics[width=1.0\linewidth]{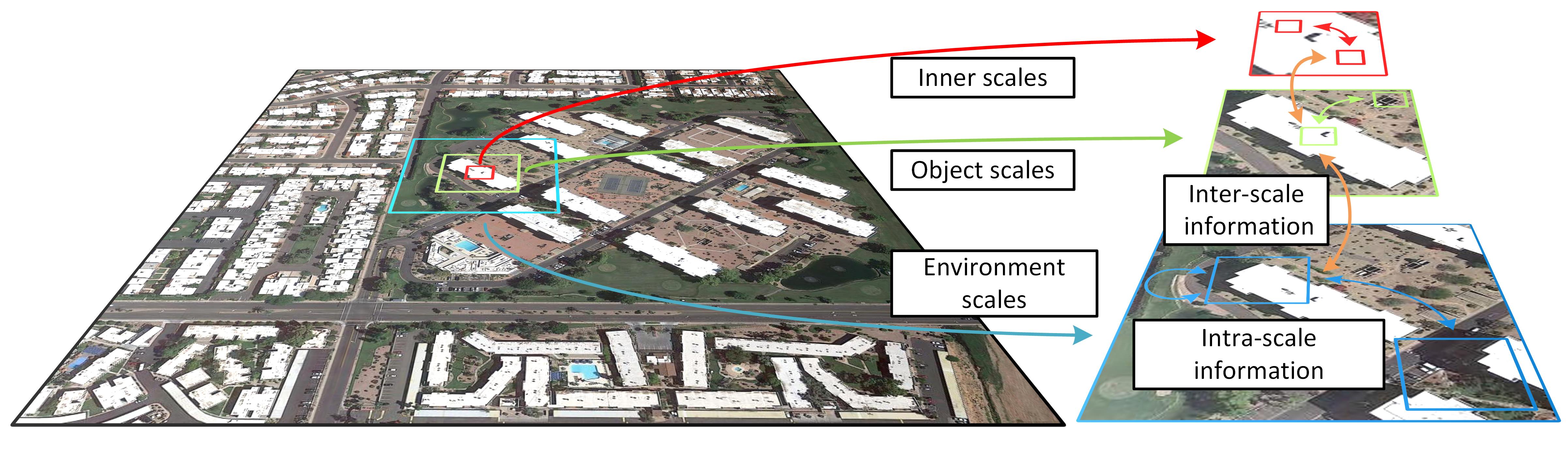}
		\caption{The shift-scale presentations in the DeepMerge.}
		\label{shitf-scale}
	\end{figure}

	As discussed above, supervised and unsupervised methods suffer from low segmentation accuracies caused by the land objects with diverse shapes and sizes and the selection of scale parameters with interpretability. To address these challenges, we propose a deep stepwise optimization method to handle segmentation in large VHR images. The proposed DeepMerge method can learn the shift-scale features (shown in  Fig.\ref{shitf-scale}) to measure the similarity between two adjacent neighbouring super-pixels. \par
	
	To comprehend the interrelations among the inner scale, object-scale, and environmental scale of objects, we introduced a shift-scale presentation module. This module can extract shift-scale features corresponding to the aforementioned scales represented in red, green, and blue as depicted in Fig.\ref{shitf-scale}. It is important to note that shift-scale differs from multi-scales, as the shift-scale pertains to object-specific scales in contrast to the fixed scales in the multi-scale. To our knowledge, this is the first effort to combine RAG with deep learning for remote sensing image segmentation. The key contributions of our study are the following:\par
	\begin{itemize}
		\item A deep learning-based region-merging method was designed to enhance large VHR image segmentation. The proposed method results in high accuracy compared to SOTA methods, correctly segmenting small and large objects.
		\item The optimal scale parameter of DeepMerge is interpretable, stabilising at 0.5, releasing the hands of scholars in selecting optimal scale parameters.
		\item To facilitate sample collection for the supervised training of DeepMerge, we designed a user-friendly graphical interface. A skilled operator can select at least 10,000 training sample pairs in a working day.
		
	\end{itemize}

	
	\section{Related works about region-merging-based segmentation}
	\label{S:2}
	Image segmentation is an essential step in remote sensing image analysis workflows. Several efforts have been made to improve the segmentation quality of remote sensing images, such as the simple linear iterative clustering method \cite{achanta2012slic}, mean-shift \cite{paris2007topological}, multi-resolution segmentation (MRS) \cite{baatz2000multi}. Contour-based methods are an important branch of segmentation methods focusing on the extraction of object boundaries \cite{martin2004learning,arbelaez2010contour,pont2016multiscale}. However, the segmentation methods that rely on a single strategy usually fail to meet the requirement of high-quality segmentation. Thus, region-merging-absed segmentation has received wide attention. In this section, we review related works concerning the rationale of RAG, optimization of scale parameters, and merging criteria.\par

	\subsection{The rationale of RAG}
\begin{figure}[h]
		\centering\includegraphics[width=0.8\linewidth]{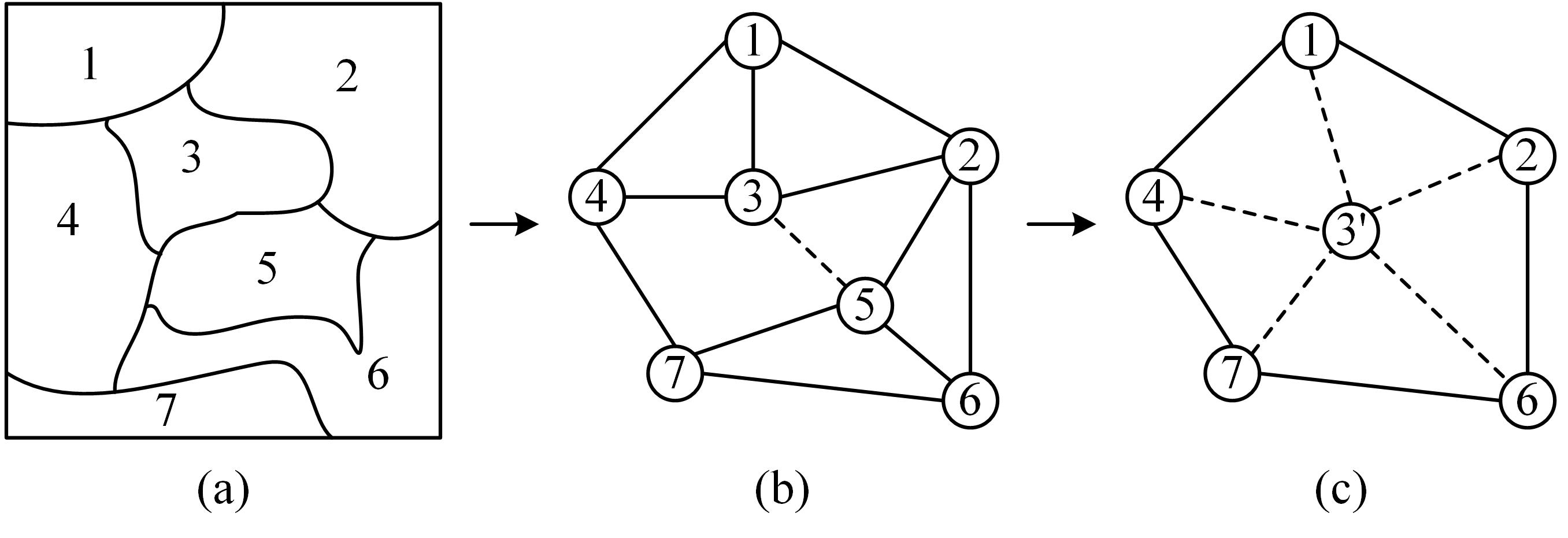}
		\caption{The construction of RAG and a region-merging step in the RAG.}
		\label{rag}
	\end{figure}

	Following the graph structure, RAG is constructed by taking the initial super-pixels as vertices and the connections between super-pixels as edges. Each vertex stores the features of the corresponding super-pixel, and each edge stores the feature distance (i.e., similarity or weight) between two vertices at its ends. Fig.\ref{rag} describes the concept of a RAG model in a hypothetical over-segmented case.\par
	
	Based on the seven over-segmented super-pixels (Fig.\ref{rag}a), a RAG (i.e., a non-directional graph) is constructed. The vertices in Fig.\ref{rag}b represent super-pixels from the original segmentation, and the edges between vertices depict the connections between super-pixels. The weight of an edge indicates the similarity between two neighbouring vertices, which is calculated by user-designed merging criteria. Supposing that the weight value of the dotted edge between \emph{$v_3$} and \emph{$v_3$} in Fig.\ref{rag}b is the smallest, indicating their high similarity, \emph{$v_3$} and \emph{$v_5$} are further merged into a new \emph{$v_{3'}$} shown in Fig.\ref{rag}c. The weights of edges connected to \emph{$v_{3'}$} are then updated. The RAG model iteratively repeats these operations until the smallest weight is higher than a user-defined scale paremeter.\par
	
	\subsection{Merging criteria in RAG}
	Merging criteria in RAG have been regarded as an important component that determines the segmentation results. To enhance the quality of segmentation, a supervised classic watershed segmentation method was improved via multispectral gradient \cite{derivaux2006watershed}. Furthermore, \citep{wassenberg2009efficient} utilized a graph-cutting heuristic method to accelerate the Minimum Spanning Tree-based algorithm. In \cite{johnson2011unsupervised}, unsupervised image segmentation and evaluation and refinement using weighted variance and Moran’s I in a series of scales was developed. The researches in \cite{zhang2014fast,zhang2013boundary,lee1991detecting,su2020machine} introduced edge strength, compactness, and the feature of standard derivation into the merging criteria. A multiscale segmentation method to achieve results through stepwise refinement guided by area and boundary features was designed \cite{chen2014optimal}. Local spectral statics were used to measure the homogeneity and heterogeneity between segments \cite{wang2019unsupervised}. To delineate dune-field landscape patches,\cite{zheng2020multiscale} proposed a new method that integrates multisource features that represent dune-field landscapes at multiple scales. The paper \cite{su2020machine} viewed the region-merging as a classification process and trained a random forest using segment-based features. However, the segment-based features failed to fully capture the details of objects, leading to the segmentation failure of complete objects.\par
	
	\subsection{Scale parameter optimisation in RAG}
	Efforts have been made to optimize scale parameters (threshold) in RAG. The segmentation output varies by setting different scale parameters in the investigated area. Low scale results in small-area segments, benefiting the segmentation of small objects; however, large objects tend to contain multiple super-pixels, leading to over-segmentation errors. The manual-optimal scale is often determined by a trial-and-error process, causing uncertainties in segmentation results \cite{lv2021improved,zhang2018object}. Therefore, scholars developed automatic and self-adaptive scale optimization methods. The majority of region-merging algorithms calculate the homogeneity in objects and the heterogeneity between objects to determine the optimal scale for objects. For example, an automated approach to parameterizing multiscale image segmentation was proposed to detect scale transitions in objects relying on the local variance \cite{druaguct2014automated}. In \cite{ming2015scale}, a spatial and spectral statics-based scale parameter selection was proposed for object-based information extraction using an average local variance graph. \cite{hu2018stepwise} developed a general stepwise evolution analysis framework for optimal scale parameter estimation using local variance and Moran's I, a measure of spatial autocorrelation. In order to better segment objects of various sizes, a scale-variable segmentation method was proposed where scale parameters are adaptively estimated \cite{zhang2014hybrid}. To obtain the objective-adaptive scale for each object, \cite{shen2019optimizing,zhang2020object} proposed object-specific optimization strategies using hierarchical tree structures of multiscale segmentation.\par

	\section{Methodology}
	\label{S:3}
	
	\subsection{Outline of the proposed method}
\begin{figure}[h]
		\centering\includegraphics[width=1.0\linewidth]{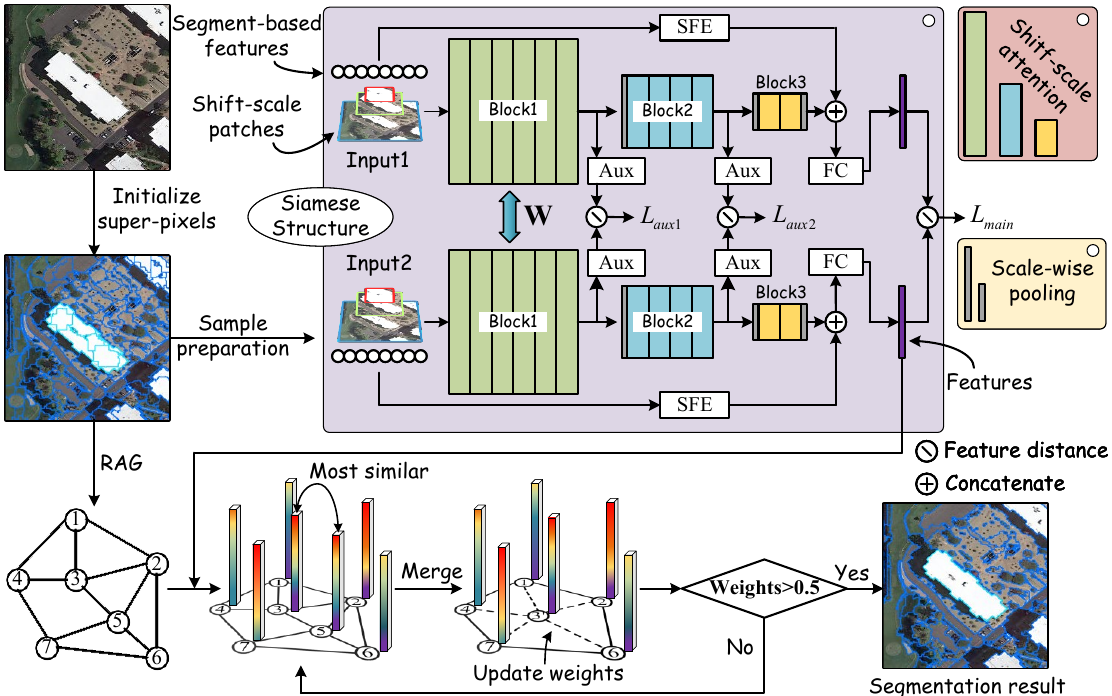}
		\caption{The workflow of DeepMerge for super-pixel segmentation in high-resolution remote sensing imagery. \emph{W} means shared weights. \emph{Aux} is the loss function auxiliary module. \emph{FC} is the fully connected layers. \emph{SFE} is the segment-absed feature embedding module.}
		\label{workflow}
	\end{figure}

	The proposed DeepMerge integrates deep learning and RAG to achieve desirable segmentation in VHR remote sensing images. Fig.\ref{workflow} summarizes the workflow of DeepMerge. The original image is firstly over-segmented into super-pixels by a standard segmentation method. Then we utilize a siamese network \cite{chopra2005learning,guo2017learning} to learn the similarity between neighbouring super-pixels. We proposed the shift-scale tansformer (S2Former) model as the backbone network. S2Former is trained using a training set composed of positive and negative samples. Pairs of adjacent super-pixels of the same object are called positive samples. On the contrary, pairs of neighbouring super-pixels of different objects are called negative samples. To train the S2Former, we manually select positive and negative samples and measure similarities (the weights in Fig.\ref{workflow}) between adjacent super-pixels. After training the S2Former, the model can measure the similarity between adjacent super-pixels. Finally, the RAG model iteratively merges the most-similar super-pixel pairs via the global best-merging strategy until all weights of edges in RAG are higher than 0.5.\par
	
	We recommend MRS for the initial (over)segmentation, a region-growing segmentation method that follows the minimum heterogeneity principle \cite{baatz2000multi}. MRS has been proven efficient in generating super-pixels as polygons in a shapefile format \cite{lv2018new}. Of course, other segmentation methods can also be used as the initial segmentation method in our framework.\par
	We describe the sample collection, shift-scale inputs, segment-based feature embedding, shift-scale attention, and the feature updating strategy for the merging process in the following sections. All codes are written either in C\# (sample selection software, shift-scale inputs extraction, RAG) or Python languages (S2Former and RAG) and tested on a computer with Windows 10 OS, an intel i7-13700K CPU (3.4GHz), 64GB RAM, and an NVIDIA GPU (RTX 4090). We open-source the code at \url{https://github.com/lvxianwei/DeepMerge}.\par
	
	\subsection{Sample collection for training S2Former}
\begin{figure}[h]
		\centering\includegraphics[width=0.78\linewidth]{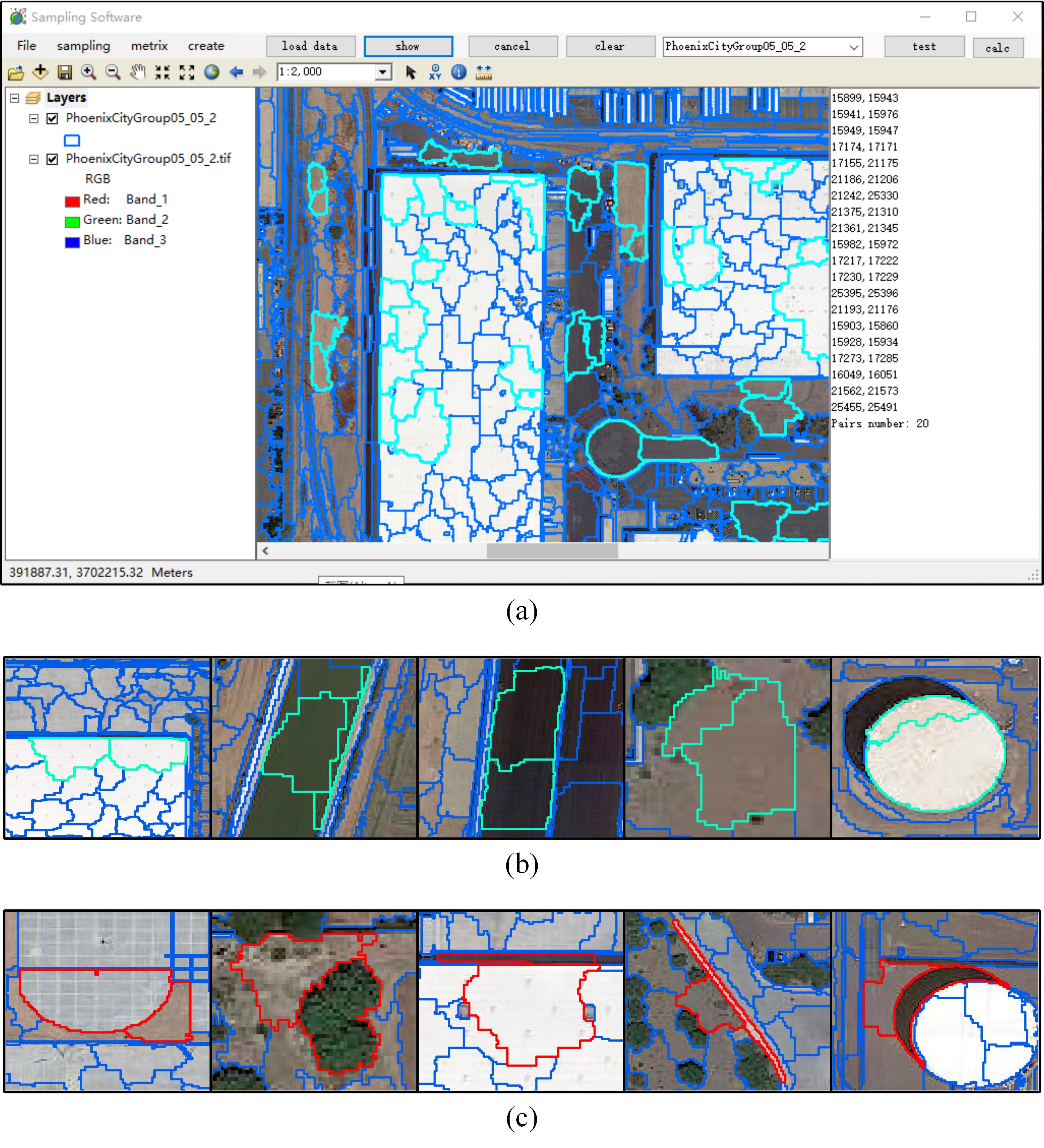}
		\caption{Sample collection graphical interface and sample pairs. (a) is the graphical interface of sample collection. (b) and (c) are the positive sample pairs and negative sample pairs, respectively.}
		\label{sample-collection}
	\end{figure}

	To train the S2Former, we manually collected samples according to the following steps. We visually analysed the super-pixels in the initial over-segmentation results, and determined the dominant categories in the study areas. Then, we selected neighbouring super-pixels (Fig.\ref{sample-collection}a) with high homogeneity serving as positive samples (Fig.\ref{sample-collection}b), and neighbouring super-pixels of different categories serving as negative samples (Fig.\ref{sample-collection}c). It is worth noting that we need to collect more negative than positive samples to account for the large variety of possible discrepancies of two different super-pixels. In VHR images, one object may contain high heterogeneous super-pixels and different adjacent super-pixels can contain similar pixels, leading to segmentation errors. Therefore, it is necessary to select many samples of both situations. In addition, we have to focus on similar neighbouring super-pixels in different categories, such as rivers and vegetations, asphalt roads and shadows, and vegetations and roads. We developed a user-friendly graphical interface for collecting the training samples, which can automatically record indexes of sample pairs selected by operators. Therefore, a skilled-operator can select at least 10,000 sample pairs in a working day, greatly improving work efficiency.\par
	
\begin{figure}[h]
		\centering\includegraphics[width=1.0\linewidth]{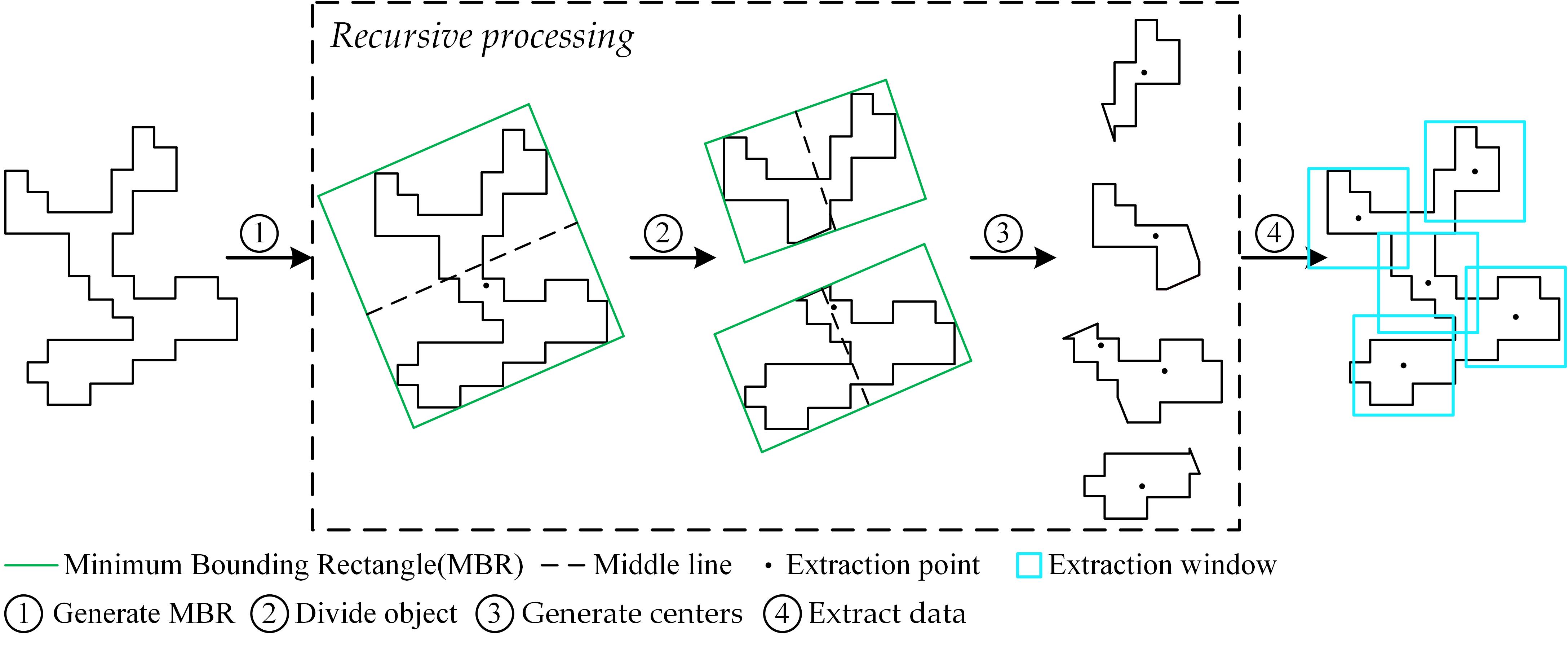}
		\caption{The basic theory about BTS.}
		\label{bts}
	\end{figure}

	The S2Former requires image patches as inputs, similar to standard deep learning models. However, the samples collected above for the proposed model are super-pixel pairs, i.e., positive and negative samples with varying shapes and sizes, as shown in Fig.\ref{sample-collection}. It leads to an input gap between super-pixel samples and the proposed model requirement. To overcome the disparity between super-pixels and the requirement for square patches, we represent super-pixels using patches with global and local information. Thus, a binary tree sampling method (BTS), as a crucial step in object-based convolutional neural networks \cite{zhang2018object,lv2019very} is implemented to generate suitable inputs that can be fed into the proposed model from super-pixel samples \cite{lv2022bts} to improve the super-pixel identification quality. The original BTS is able to partition a super-pixel into sub-super-pixels by recursively dividing the super-pixel into two parts until a user-defined threshold is reached. Based on the positions of sub-parts, shift-scale data can be extracted in an adaptive window size. Fig.\ref{bts} demonstrates how BTS works. For more information, please refer to \cite{lv2022bts}.\par
	
	The number of the threshold and the extraction window size are the key issues in the sampling procedure. Thus, we improve BTS by developing an automatic strategy. According to \cite{lv2022bts}, up to three positions are needed to extract information for the representation of a super-pixel. The window size is determined by the ratio of the intersection of the windows and the super-pixels. Therefore, each super-pixel corresponds to different window sizes. In order to obtain local and global information of super-pixels, we designed extraction windows in three scales (\emph{${P}_{1}$}, \emph{${P}_{2}$}, and \emph{${P}_{3}$}). The automatic strategy is summarized as follows.\par
	
	\IncMargin{1em}
	\begin{algorithm} \SetKwData{Left}{left}\SetKwData{This}{this}\SetKwData{Up}{up} \SetKwFunction{Union}{Union}\SetKwFunction{FindCompress}{FindCompress} \SetKwInOut{Input}{input}\SetKwInOut{Output}{output}
		\Input{A super-pixel \emph{Seg} and a centre position \emph{Pos} for extracting image patches} 
		\Output{Three patches \emph{${P}_{1}$}, \emph{${P}_{2}$}, and \emph{${P}_{3}$} in four spatial scales}
		\BlankLine 
		\emph{patch} $\leftarrow$ ExtractPatch(\emph{Pos}, 5)\; 
		\emph{iter} $\leftarrow$ 0 \;
		\While{\emph{${P}_{1}$},\emph{${P}_{2}$}== $null$ }{ 
			\emph{$intersect\_area$} $\leftarrow$ \emph{Seg}.Intersect(\emph{patch}).area\; 
			\emph{ratio} $\leftarrow$ \emph{$intersect\_area$} / \emph{patch}.area\;
			\If(\emph{${P}_{1}$} $\leftarrow$ \emph{patch};)
			{\emph{ratio} $<$= 0.90 and \emph{${P}_{1}$} == $null$}{\label{lt} } 
			\If(\emph{${P}_{2}$} $\leftarrow$ \emph{patch};)
			{\emph{ratio} $<$= 0.30 and \emph{${P}_{2}$} == $null$}{\label{lt} } 
			\emph{iter}$\leftarrow$\emph{iter}+1\;
			\emph{patch} $\leftarrow$ ExtractPatch(\emph{Pos}, 5 + \emph{iter} $\times$ 5)\;  
		} 
		\emph{${P}_{3}$} $\leftarrow$ ExtractPatch(\emph{Pos}, \emph{${P}_{2}$}.width + (\emph{${P}_{2}$}.width – \emph{${P}_{1}$}.width)) \;
		
		\textbf{final}\;
		\textbf{return} \emph{${P}_{1}$}, \emph{${P}_{2}$}, and \emph{${P}_{3}$}\;
		\caption{shift-scale input data generation}
		\label{multi-level data} 
	\end{algorithm}
	\DecMargin{1em} 
	
	\emph{Seg} is a super-pixel to be presented by multiple patches in DeepMerge. The red stars in \emph{Seg} are the positions for extracting square patches (i.e., the red middle star in Fig.5 is the \emph{Pos}). \emph{ExtractPatch}(.) is a function applied to extract a square \emph{patch} with the \emph{Pos} as the centre and the initial side length of 5 pixels. The intersection area is calculated by the function \emph{Intersect}(.) between \emph{Seg} and \emph{patch} is \emph{intersect\_area}. We iteratively increase the width of the square patch by 5 pixels. The \emph{intersect\_area} will increase accordingly. Meanwhile, the area \emph{ratio} of the intersection to the current patch will decrease from 100\%. When the \emph{ratio} is firstly lower than 90\%, the current patch will serve as \emph{${P}_{1}$}. When the \emph{ratio} is smaller than 30\%, the current patch will serve as \emph{${P}_{2}$}. Thus, \emph{${P}_{3}$} can define by \emph{Pos} and width difference of \emph{${P}_{1}$} and \emph{${P}_{2}$} shown in the above pseudo codes. The 90\% and 30\% are defined by the human visual observation of the objects. We found that when the \emph{ratio} approaches 90\%, the current patch can capture most inner information without much external information. The balance information on inner and external objects can be captured by the current patch when the ratio approaches to 30\%. Note that other positions in the super-pixel share the same multi-scale window sizes. Because the super-pixels are of different shapes, the square patches extracted from each super-pixel are also different.
	
	
	As most super-pixels tend to be over-segmented, an object can contain multiple super-pixels. Thus, information from the \emph{${P}_{1}$} and \emph{${P}_{2}$} can sometimes only represent the object partially. To enhance the representation of global objects, information from the \emph{${P}_{3}$} needs to be extracted. In certain cases, some small super-pixels are completely segmented without over-segmentation errors. In this situation, the designed shift-scale information strategy provides not only global information for better overall representation but also neighbouring information that improves the robustness of region merging. Thus, small and large super-pixels can be well-represented following the designed protocol.

	\subsection{S2Former model}
\begin{figure}[h]
		\centering\includegraphics[width=1\linewidth]{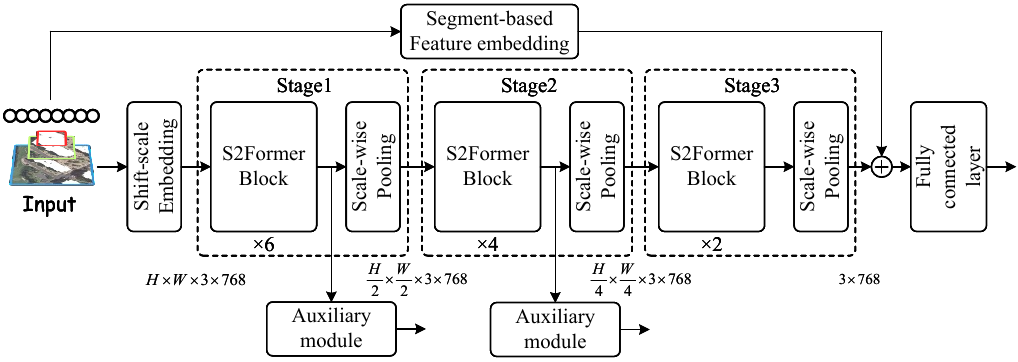}
		\caption{The architecture of the S2Former in the DeepMerge.}
		\label{s2former}
	\end{figure}

	To learn the similarity between neighbouring super-pixels, we propose an S2Former model as the backbone in the Siamese network, which is used for supervised contrastive learning \cite{guo2017learning}. The basic structure of the Siamese network is presented in Fig.\ref{workflow}. The features of negative and positive samples can be extracted by a weight-shared backbone. The similarity is obtained by the loss function that calculates the distance in the feature space. The S2Former comprises shift-scale attention modules, scale-wise pooling, auxiliary modules, and segment-based feature embedding modules shown in Fig.\ref{s2former}. The mentioned-above basic component modules in the S2Former are described in the following sections.\par
	
	\subsubsection{Shift-scale attention mechanism}
\begin{figure}[h]
		\centering\includegraphics[width=1.0\linewidth]{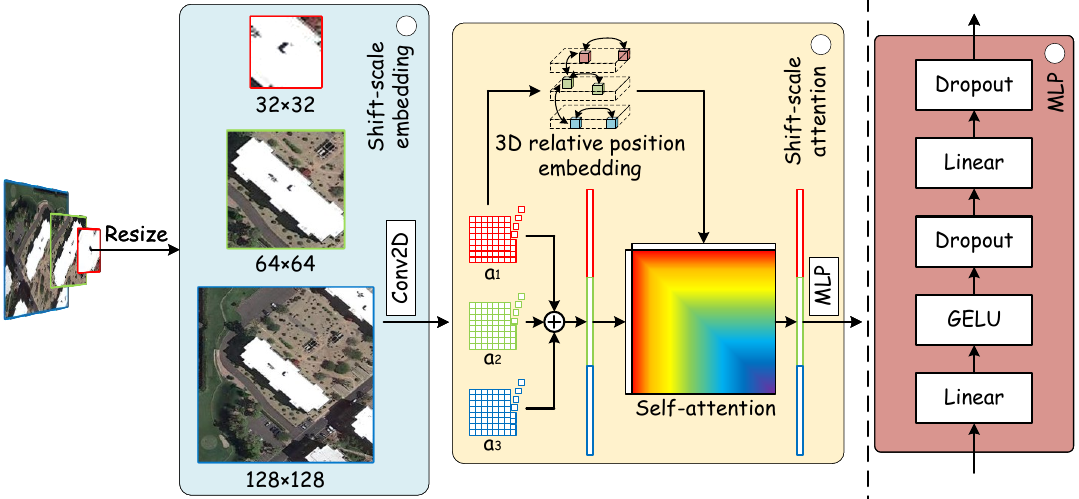}
		\caption{Shift-scale attention mechanism.}
		\label{shift-scale-attention}
	\end{figure}

	Shift-scale attention forms the basic S2Former block, which can capture long-range and cross-scale pixel dependencies based on the self-attention mechanism with 3D relative position embedding (Fig.\ref{shift-scale-attention}). Inspired by human visual concertation, the attention mechanism focuses attention on important information, thereby saving resources and extracting accurate information in a rapid manner\cite{arnab2021vivit}. The S2Former is a feature extraction structure composed of multi-head modules using a shift-scale attention mechanism. The input patches firstly are resized into 32$\times$32, 64$\times$64, and 128$\times$128, then embedded into three 8$\times$8 vectors of features via shift-scale embedding moudle \emph{Conv2D} (2D convolution functions), whose kernel sizes are 4$\times$4, 8$\times$8, and 16$\times$16, respectively. The embedded features from shift-scale embedding are then flattened into 1D vectors and concatenated for shift-scale attention modules. The \emph{Linear} is a one-layer fully connected network. MLP denotes the Multilayer Perception, and the Gaussian error linear units (GELU) is the activation function \cite{hendrycks2016gaussian}. The layer \emph{Dropout} is employed as a regularization strategy.\par
	
	Eq.\ref{attention} depicts the process of the attention mechanism. \emph{V}(=\emph{$AW^v$}) is the output of the input \emph{A}\{$a_1,a_2,a_3$\} (Fig.\ref{shift-scale-attention}) linear transformed by the weight \emph{$W^v$}. In visual image processing, \emph{A} is a 1D vector composed of embedded features $a_1$, $a_2$, and $a_3$. The \emph{softmax} function plays the role of scorer for \emph{V} based on query (\emph{Q} = \emph{$AW^q$}) and key (\emph{K}=\emph{$AW^k$}), where \emph{$W^q$}, \emph{$W^k$} are the weights for updating in the backpropagating process, \emph{Q} and \emph{K} are two special matrices used for searching the importance of pixels. Inspired by the 2D relative-position embedding, we expand it to a 3D relative-position embedding to measure the relative spatial relationships \emph{B} of the shift-scale inputs. The features of the self-attention output are finally fed into an MLP. \par
	
	\begin{equation}
	Attention(Q,K,V)=softmax\left( \frac{Q{{K}^{T}}}{\sqrt{{{d}_{k}}}} +B \right)V
	\label{attention}
	\end{equation}
	
	\begin{equation}
	softmax {{\left( z_{i} \right)}}=\frac{\exp \left( {{z}_{i}} \right)}{\sum\nolimits_{j=1}^{K}{\exp \left( {{z}_{j}} \right)}}
	\label{softmax}
	\end{equation}
	where \emph{$d_k$} is the dimension of \emph{K}. The output of the attention module, named head, contains scored information in \emph{A}. The above process is the basic principle of the shift-scale attention mechanism. The \emph{softmax} function applies the standard exponential function to each element ${z}_{i}$ of the input vector \emph{z} and normalizes these values by dividing by the sum of all these exponentials; this normalization ensures that the sum of the components of the output vector \emph{z} is 1. The shift-scale attention forms the S2Former block in Fig.\ref{s2former}.\par

	\subsubsection{Scale-wise pooling}
	
\begin{figure}[h]
		\centering\includegraphics[width=0.8\linewidth]{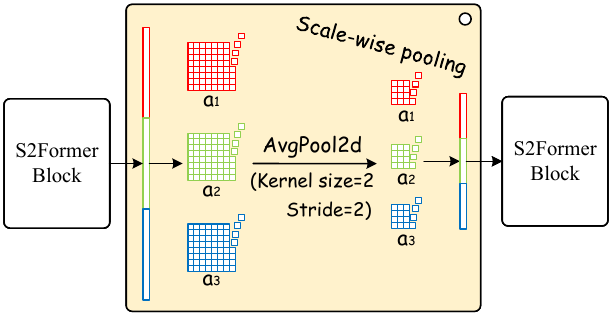}
		\caption{Scale-wise pooling module.}
		\label{scale-wise-pooling}
	\end{figure}

	To decrease the model size, we propose a scale-wise pooling which is able to pool the outputs of S2Former block in three scales. Fig.\ref{scale-wise-pooling} depicts the process of scale-wise pooling. The output of a S2Former block is a 1D vector composed of features in three scales marked in red, green, and blue. The featuers are firstly un-flattened into the 2D state in three scales, respectively. The average pooling is then applied to reduce the size of 2D features. Finally, the new 2D features in three scales are flattened and concatenated into a 1D vector, whose size is reduced by half. 
	
	\subsubsection{Auxiliary module and segment-based feature embedding}
\begin{figure}[h]
		\centering\includegraphics[width=1.0\linewidth]{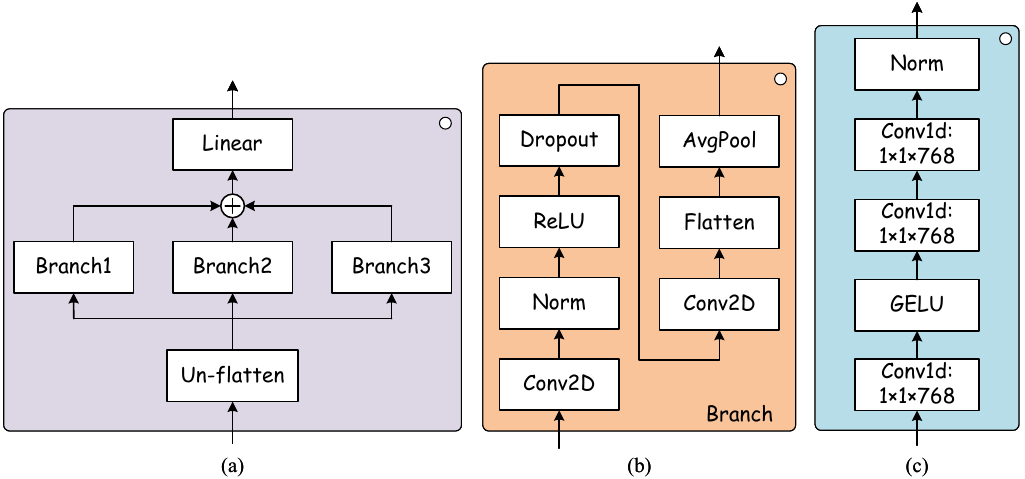}
		\caption{Auxiliary module. (a) is the auxiliary module; (b) is the branch of the auxiliary module; (c) is the segment-based feature embedding.}
		\label{auxiliary-module}
	\end{figure}

	The auxiliary module can assist in improving the performance and robustness of the S2Former. The 1D features of the S2Former block are processed individually in three branches related to the three scales shown in Fig.\ref{auxiliary-module}a and Fig.\ref{auxiliary-module}b. The \emph{Conv2D} (Fig.\ref{auxiliary-module}b) and \emph{Conv1D} (Fig.\ref{auxiliary-module}c) are two-dimensional and one-dimensional convolution functions, respectively. The \emph{Norm} block is the batch normalization for inputs. The \emph{ReLU} is the Linear rectification function. The \emph{AvgPool} is a global pooling layer.

	The segment-based feature embedding block can encode engineered features to integrate other deep features extracted by networks (Fig.\ref{auxiliary-module}c). A total of eighteen features are designed for segment-based feature embedding module, including texture features, statistical features, shape features, standard deviation of each band, mean value of each band, shape indicator, compactness, brightness, border indicator, and resizing factors. These features are computed considering the pixels within the segment. They capture different characteristics to assess region similarities with respect to deep learning features, which are calculated using a fixed window size. Their calculations follow:
	
	\begin{equation}
	Mea{{n}_{i}}=\frac{1}{n}\sum\limits_{j=1}^{n}{{{v}_{j}}}
	\label{mean}
	\end{equation}
	\begin{equation}
	St{{d}_{i}}=\sqrt{\frac{1}{n-1}\sum\limits_{j=1}^{n}{{{\left( {{v}_{j}}-Mea{{n}_{i}} \right)}^{2}}}}
	\label{std}
	\end{equation}
	\begin{equation}
	Shape=\frac{l}{4\sqrt{C}}
	\label{shape}
	\end{equation}
	\begin{equation}
	Compactness=l\sqrt{n}
	\label{compactness}
	\end{equation}
	\begin{equation}
	Brightness=\frac{1}{{{w}^{B}}}\sum\limits_{i=1}^{K}{w_{i}^{B}Mea{{n}_{i}}}
	\label{brightness}
	\end{equation}
	\begin{equation}
	Border=\frac{l}{2\left( length+width \right)}
	\label{border}
	\end{equation}
	where \emph{i} (\emph{i}=1,2,3) indicates the \emph{i}th band, \emph{n} denotes the super-pixel size in pixels, ${v}_{j}$ denotes the \emph{j}th pixel value, and ${Mean}_{i}$ denotes the mean value of the \emph{i}th band in a super-pixel. The ${Std}_{i}$ denotes the standard deviation of the \emph{i}th band in a super-pixel. The Shape denotes the shape indicator defined by the perimeter (\emph{l}) of a super-pixel and the perimeter (\emph{C}) of the minimum bounding rectangle (\emph{MBR}) of a super-pixel. The \emph{length} and \emph{width} are the long edge and the short edge of the \emph{MBR}. The \emph{length} and \emph{width} are used to define the border indicator (\emph{Border}) and \emph{Compactness}. \emph{K} denotes the number of bands, and $w_{i}^{B}$ denotes the related weight.\par
	
	\subsubsection{Loss function}
	\label{loss-function}
	The output of the S2Former is a 1D vector storing features used for super-pixel representation, and it will be used in calculating similarity in an RAG model. The features of positive sample pairs or negative sample pairs are extracted by the network and are supervised by the loss function. Given a pair of super-pixels, The \emph{$L_{aux1}$}, \emph{$L_{aux2}$}, and \emph{$L_{main}$} in the Fig.\ref{workflow} are calculated by the loss function:\par
	
	\begin{equation}
	Loss=\alpha {{\left\| A_{i}^{left}-A_{i}^{right} \right\|}_{2}}+\left( 1-\alpha  \right)max\left( 0,\left( \lambda -{{\left\| A_{i}^{left}-A_{i}^{right} \right\|}_{2}} \right) \right)
	\label{loss}
	\end{equation}
	where \emph{A} means the feature vector in a super-pixel, \emph{i} indicates the feature item from the \emph{A}. $\alpha$ is a binary indicator (positive pair: $\alpha$=1 ; negative pair: $\alpha$=0). $\lambda$ is a user-defined parameter that represents the cluster centre of the Euclidean distance in feature space between negative sample pairs (we recommend $\lambda$=1). The function \emph{max}(.,.) can return the maximum value from two parameters. In the proposed DeepMerge, the distance in feature space between positive sample pairs should be close to 0, while the distance between negative sample pairs should be close to 1 if less than 1, which predicts that the optimal scale parameter of the proposed method will be 0.5. Once the super-pixel features are extracted by the S2Former, they further serve as the vertex features in the RAG model (described in Fig.\ref{workflow}). The final loss of the whole S2Former is calculated by the sum of weighted loss values above. \par
	
	\begin{equation}
	L_{final}= L_{main} + 0.1 \times L_{aux1} + 0.2 \times L_{aux2}
	\label{final_loss}
	\end{equation}
	
	\subsection{Merging criteria and feature updating}
	
	Merging criteria are designed for calculating the similarity between super-pixels, i.e., edge weight in the RAG model. We use the Euclidean distance of features from two neighbouring super-pixels as the merging criteria in the proposed method:\par
	
	\begin{equation}
	MC={{\left\| A_{i}^{left}-A_{i}^{right} \right\|}_{2}}
	\label{mc}
	\end{equation}
	where \emph{MC} is the Euclidean distance of two group features, which is the similarity of two super-pixels, recorded as the weight of the connected edge in the RAG model. Conventional merging criteria updates edge weights of a newly merged super-pixel by re-extracting features and re-calculating weights between the super-pixel and its adjacent super-pixels. However, deep features of a newly merged super-pixel can cause inefficiency in region-merging if they are re-extracted by the pre-trained model. In the Deeop-SO, the features of a new super-pixel are calculated via the weighted average of the original two features by Eq.\ref{updatemc}:\par
	
	\begin{equation}
	A_{i}^{left+right}=\frac{1}{m+n}\left( mA_{i}^{left}+nA_{i}^{right} \right)
	\label{updatemc}
	\end{equation}
	where $A_{i}^{left+right}$ are the features of new super-pixels. \emph{m} and \emph{n} denote the feature vector weights of the super-pixel \emph{lfet} and \emph{right}. \emph{left} and \emph{right} are the number of extracting centres by BTS in the super-pixel \emph{left} and \emph{right}. Therefore, the feature vector weight of the new super-pixel is \emph{m}+\emph{n}. In fact, calculating the features of the newly merged super-pixel is calculating the spatial clustering center of the features of all intial super-pixels it contains. The feature vector of a initial super-pixel is the average of features extracted from multiple extracting centres by BTS.\par
	
	\subsection{Segmentation accuracy estimation}
	
	To assess the segmentation performance of the proposed DeepMerge, two groups of accuracy assessment metrics are applied, considering over-segmentation, under-segmentation, and whole-segmentation performance. The first group of metrics includes \emph{precision}, \emph{recall}, and \emph{F} value  \cite{zhang2015segmentation}. The second group of metrics includes the global over-segmentation error (\emph{GOSE}), global under-segmentation error (\emph{GUSE}), and total error (\emph{TE}) \cite{su2017local}. These measurement metrics require polygon segmentation results and vectorized reference objects, which have been proven to be effective and robust for measuring the local and global segmentation performance from various aspects. The calculations of these metrics are presented in Table \ref{Tab1}. \emph{S} is the set of polygon segmentation results containing \emph{M} segments \{${S}_{1}$, ${S}_{2}$, …, ${S}_{M}$\}, and \emph{R} is the set of polygon reference objects containing \emph{N} reference objects \{${R}_{1}$, ${R}_{2}$, …, ${R}_{N}$\}. $\left|*\right|$ represents the area of a segment. ${R}_{i,max}$ denotes the largest area reference object related to the segment ${R}_{i}$, and ${R}_{i,max}$ denotes the largest area segment related to the reference object ${R}_{i}$. ${R}_{ij}$ denotes the set of segments related to ${R}_{i}$. ${S}_{i}\cap{R}_{i,max}$ is the intersection of ${S}_{i}$ and ${R}_{i,max}$, and the ${S}_{i}\cup{R}_{i,max}$ the union of them. The difference set $\left|{R}_{i}\backslash {S}_{i,max}\right|$ contains pixels in ${R}_{i}$, but not in ${S}_{i,max}$. $\alpha$ is set as 0.5. The $\uparrow$ means higher values with better performance and vice versa for the $\downarrow$.\par
	
	\begin{table}[h]
		\centering
		\caption{Assessment metrics used for segmentation accuracy estimation.}
		\resizebox{\linewidth}{!}{
			\begin{tabular}{l l l l}
				\hline
				\textbf{\makecell[c]{Assessment metrics}} & \textbf{\makecell[c]{Formulas}}  & \textbf{\makecell[c]{Range}} & \textbf{\makecell[c]{Trend}}\\
				\hline
				\makecell[c]{\emph{precision}}   & \makecell[c] {$\left| S \right|=\sum\limits_{i=1}^{M}{\left| {{S}_{i}} \right|}$ \\  
					$precision=\sum\limits_{i=1}^{M}{\left| {{S}_{i}}\bigcap {{R}_{i,\max }} \right|}/\left| S \right|$ \\
					$\left| R \right|=\sum\limits_{i=1}^{N}{\left| {{R}_{_{i}}} \right|}$ } 
				& \makecell[c]{[0,1]} 
				& \makecell[c]{$\uparrow$}\\
				\makecell[c]{\emph{recall}}	    & \makecell[c]{$\left| R \right|=\sum\limits_{i=1}^{N}{\left| {{R}_{_{i}}} \right|}$\\
					$recall=\sum\limits_{i=1}^{N}{\left| {{R}_{i}}\bigcap {{S}_{i,\max }} \right|}/\left| R \right|$}
				& \makecell[c]{[0,1]}
				&\makecell[c]{$\uparrow$}\\
				\makecell[c]{\emph{F}}   & \makecell[c]{${F}={1}/{\left( \alpha \frac{1}{p}+(1-\alpha )\frac{1}{r} \right)}\;$}
				& \makecell[c]{[0,1]}
				& \makecell[c]{$\uparrow$}\\
				\makecell[c]{$GOSE$}	    &\makecell[c]{
					$GOSE=\frac{1}{\left| R \right|}\sum\limits_{i=1}^{N}{\left| {{R}_{i}} \right|\frac{\left| {{R}_{i}}\backslash {{S}_{i,\max }} \right|}{\left| {{R}_{_{i}}} \right|-1}}$}
				& \makecell[c]{[0,1]}
				& \makecell[c]{$\downarrow$}\\
				\makecell[c]{$GUSE$}     & \makecell[c]{
					$GUSE=\frac{1}{\left| R \right|}\sum\limits_{i=1}^{N}{\min \left( \left| {{R}_{i}}\bigcup {{S}_{ij}} \right|\backslash \left| {{R}_{i}}\bigcap {{S}_{ij}} \right|,\left| {{R}_{i}} \right| \right)}$}
				& \makecell[c]{[0,1]} & \makecell[c]{$\downarrow$}\\
				\makecell[c]{$TE$} &\makecell[c]{ $TE=GOSE+GUSE$ }
				& \makecell[c]{[0,2]} &\makecell[c]{$\downarrow$}\\
				\hline
			\end{tabular}
		}
		
		\label{Tab1}
	\end{table}
	
	\section{Experimental Results}
	\label{S:4}
	
	\subsection{Dataset}
	
\begin{figure}[h]
		\centering\includegraphics[width=1.0\linewidth]{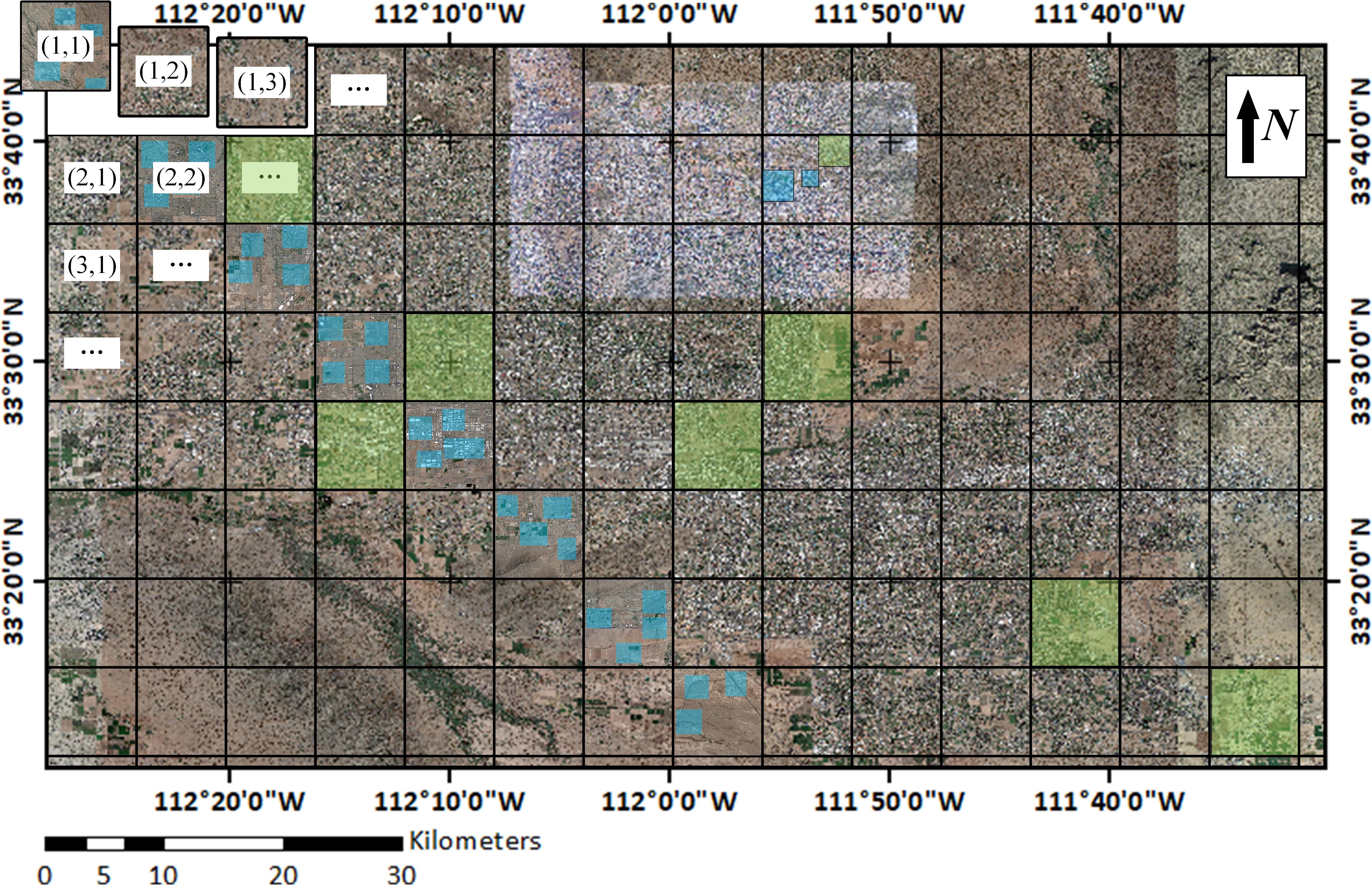}
		\caption{The study area of the Phoneix city cluster. The translucent blue tiles are used for training and the green tiles are used for accuracy assessment.}
		\label{dataset}
	\end{figure}

	The dataset used in this study covers nine cities that belong to the Phoenix city cluster, Arizona, U.S., including Phoenix, Glendale, Scottsdale, Tempe, Mesa, Chandler, Peoria, Surprise, and Goodyear. The images are composites from Google Earth \cite{yu2012google} with a variety of sensors (e.g., WorldView, QuickBird, IKONOS, etc.) captured at different times. The image contains 18.7B pixels ($182,272\times102,626$, covering 5,660 ${km}^{2}$) with 0.55-meter resolution and RGB bands encoded as 8-bit integral values. A variety of scenes are included, e.g., urban residential zones, urban green spaces, industrial zones, rural farmlands, water areas, and bare lands (Fig.\ref{dataset}).\par
	
	We partition the image into a total of 135 tiles encoded by row and column numbers (Fig.\ref{dataset}). The average size of a tile is 167M ($12,800\times12,800$) pixels. The training of the proposed DeepMerge requires negative and positive samples. In the Phoenix dataset, negative and positive pairs are selected in the patches with blue masks, as shown in Fig.\ref{dataset}. A total of 71,948 super-pixel pairs (54,945 negative sample pairs and 17,003 positive sample pairs) involving 92,421 super-pixels are manually collected as training data. There is a total of 47,995,928 super-pixels in the whole study area. The training data accounts for 0.19\% (=92,421/47,995,928) of the total number of super-pixels.\par
	To assess the performance of DeepMerge, a total of 3,776 polygons are manually digitized as reference objects. To validate the transferability of DeepMerge, the reference objects, covering a variety of land uses (Fig.\ref{testdata}), are selected in the green mask tiles in Fig.\ref{dataset}.\par
	\begin{figure}[h]
		\centering\includegraphics[width=1.0\linewidth]{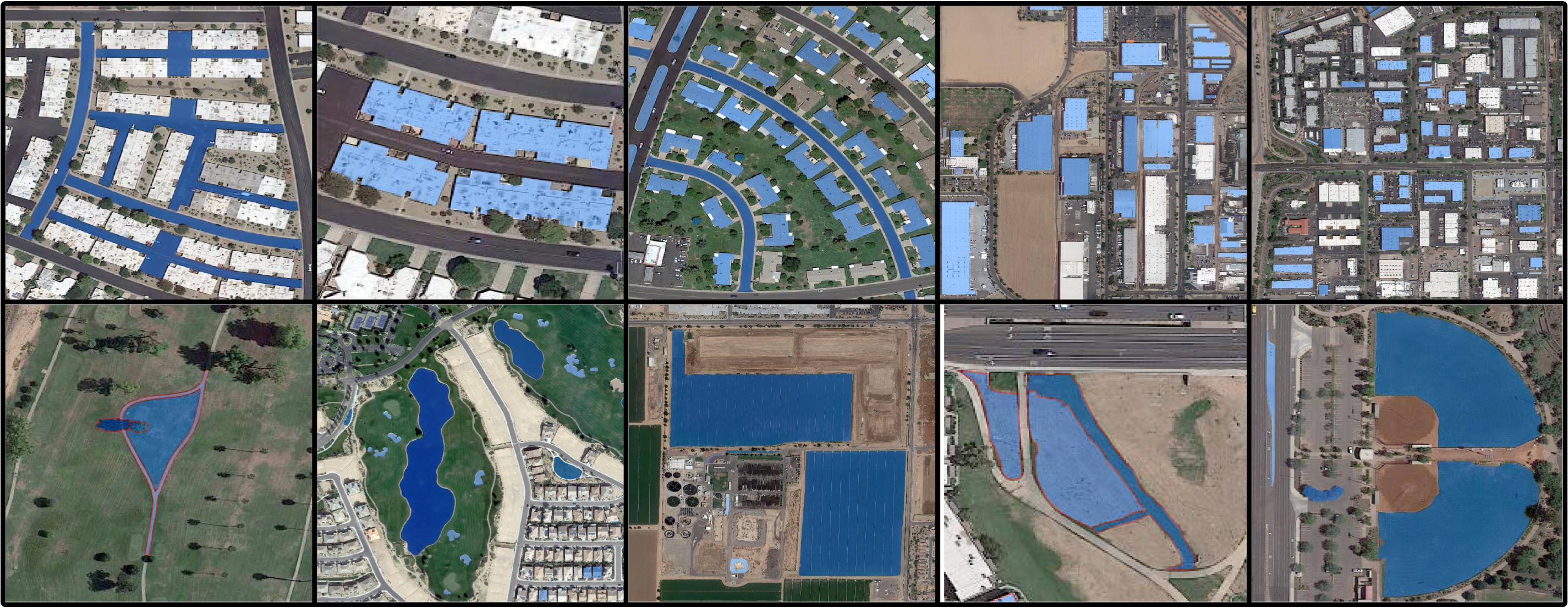}
		\caption{Examples of reference objects.}
		\label{testdata}
	\end{figure}
	
	\subsection{Region-merging-based image segmentation results}
	
	For consistency, we set the shape, compactness, and scale parameters of MRS for initializing over-segmentation as 0.5, 0.5, and 25. Too many segments can lead to low efficiency of region-merging. The MRS with these super-pixels applied to segment minimum objects in the dataset with little over-segmentation errors, maintaining a balance between over-segmentation errors and reducing super-pixel number. Relying on the same initial over-segmentation and the same RAG model, ten bottom-to-top supervised, and unsupervised methods are selected as competing algorithms, including BCMS \cite{zhang2013boundary}, FHS \cite{zhang2014fast}, HRM \cite{zhang2014hybrid}, CSVD \cite{chen2015image}, USIH \cite{hu2017unsupervised}, Local-SA \cite{yang2017region}, OMS \cite{shen2019optimizing}, OSO(fine and coarse) \cite{zhang2020object}, MLRM \cite{su2020machine}, IOseg \cite{lv2021improved}. These methods differ in optimization strategies, including merging criteria (e.g., edge penalty), object optimization, scale sets, and supervised segmentation. Diverse merging criteria suggest different scale parameters. Some multiscale segmentation methods (BCMS and FHS) set scale parameters by trial and error. The scale parameters in HRM, CSVD, Local-SA, MLRM, and IOseg are set as recommended. USIH, OMS, and OSO automatically generate specific objects with optimized scales, where OSO can produce coarse and fine segmentations at the same time. In our experiments, we set optimal scale value of DeepMerge to 0.5. In addition to standard image segmentation methods, deep learning-based semantic segmentation has made great progress in past years. The decoder-encoder neural networks are able to predict pixel-wise labels in the image. Novel modules have been designed to be applied in various applications to improve semantic segmentation accuracy. Standard deep learning based semantic segmentation methods, UNet \cite{unet}, UNet++ \cite{unet++}, U2Net \cite{u2net}, UNetFormer \cite{unetformer}, FCN \cite{fcn}, SegNet \cite{segnet}, DeepLab \cite{deeplabv1}, PspNet \cite{pspnet}, ABCNet \cite{abcnet}, MAResUNet \cite{mresunet}, EANet \cite{eanet}, CCnet \cite{ccent}, SegFormer \cite{segformer}, DenseASPP \cite{denseaspp}, and ENet \cite{enet} are selected as the competing algorithms. We labeled the blue mask areas (Fig.\ref{dataset}) as the training dataset for semantic segmentation networks. We preserved the boundaries of semantic segmentation result in the green mask areas serving as polygon objects in comparison to DeepMerge. The details about labeling dataset, training networks, and outputing image segmentation are shown in \ref{details-for-e2e}.  The \emph{F} values and \emph{TE} vales of segmentation results based on the semantic segmentation methods are low and high, respectively, caused by low \emph{precision} and high \emph{GUSE}, meaning serious under-segmentation errors in the results caused by the pixel adhesion, whose details are described in \ref{details-for-e2e}.\par
	
	The segmentation performances of DeepMerge and other competing methods are shown in Table \ref{results}. A sensitivity analysis of the scale parameter is reported in Section \ref{scale-analysis}. The \emph{precision}(0.9772) of the segmented results from MRS, the initial segmentation approach, is the highest, and the \emph{GUSE} value (0.0346) of MRS is the lowest among all investigated methods. These two metrics are closely related to over-segmentation errors. In general, higher \emph{precision} and lower \emph{GUSE} values indicate stronger over-segmentation errors and weaker under-segmentation errors. We notice that the proposed DeepMerge achieves the best performance in the rest metrics compared with other methods. The \emph{F} value of DeepMerge is 0.9550 higher than FHS, the second-best method (\emph{F} value: 0.8465). The \emph{TE} values (0.0895) of the DeepMerge are the lowest, suggesting small segmentation errors. Note that the proposed DeepMerge, among all competing algorithms, achieves the highest \emph{recall} value and the lowest \emph{GOSE} value at the same time, indicating its superiority and robustness. The image segmentation results preserved from semantic segmentations show weak performance compared to standard superpixel segmentation methods. The \emph{precision} and \emph{GOSE} values of these methods are extremely low, as similar nearby objects can be clustered into the same object, resulting in substantial undersegmentation errors.\par

	\begin{table}[h]\scriptsize 
		\centering
		\caption{Image segmentation performance of the DeepMerge and other competing methods.}
		\begin{tabular}{l l l l l l l l l l}
			\hline
			\textbf{\makecell[c]{Method}}
			& \textbf{\makecell[c]{precision$\uparrow$}}
			& \textbf{\makecell[c]{recall$\uparrow$}} 
			& \textbf{\makecell[c]{F$\uparrow$}}
			& \textbf{\makecell[c]{GOSE$\downarrow$}}
			& \textbf{\makecell[c]{GUSE$\downarrow$}}
			& \textbf{\makecell[c]{TE$\downarrow$}}\\
			\hline
			\makecell[c]{MRS}	 
			& \textbf{0.9772} & 0.1913 & 0.3200 
			& 0.1028 & \textbf{0.0346} & 0.1374  
			\\
			\hline
			\makecell[c]{BCMS}	 
			& 0.9540 & 0.4233 & 0.5864 
			& 0.1392 & 0.0431 & 0.1823 
			\\
			\makecell[c]{FHS} 
			& 0.9568 & 0.7590 & 0.8465 
			& 0.1247 & 0.0459 & 0.1706 
			\\
			\makecell[c]{HRM} 
			& 0.9714 & 0.3288 & 0.4913 
			& 0.1246 & 0.0394 & 0.1641 
			\\
			\makecell[c]{CSVD} 	 
			& 0.9657 & 0.5954 & 0.7366 
			& 0.1762 & 0.0504 & 0.2265 
			\\
			\makecell[c]{USIH} 	 
			& 0.8870 & 0.6504 & 0.7505 
			& 0.1164 & 0.0483 & 0.1647 
			\\
			
			\makecell[c]{Local-SA}   
			& 0.7062 & 0.8179 & 0.7580 
			& 0.0610 & 0.1370 & 0.1980 
			\\
			
			\makecell[c]{OMS}   
			& 0.9733 & 0.3154 & 0.4764 
			& 0.1145 & 0.0819 & 0.1964 
			\\
			
			\makecell[c]{OSO(fine)}   
			& 0.9700 & 0.4624 & 0.6262 
			& 0.1387 & 0.0406 & 0.1793 
			\\
			
			\makecell[c]{OSO(course)} 
			& 0.9586 & 0.5150 & 0.6700 
			& 0.1283 & 0.0469 & 0.1751 
			\\
			
			\makecell[c]{MLRM} 	
			& 0.9699 & 0.3860 & 0.5522 
			& 0.1606 & 0.0390 & 0.1996 
			\\
			
			\makecell[c]{IOseg} 	
			& 0.7346 & 0.8737 & 0.7981 
			& 0.0903 & 0.0563 & 0.1466 
			\\
			\hline
			
			\makecell[c]{UNet} 	
			& 0.2424 & 0.9165 & 0.3834 
			& 0.0558 & 0.4076 & 0.4634 
			\\
			
			\makecell[c]{FCN} 	
			& 0.0364 & 0.9029 & 0.0700 
			& 0.0615 & 0.4468 & 0.5083
			\\

			\makecell[c]{SegNet} 	
			& 0.0678 & 0.9222 & 0.1263 
			& 0.0470 & 0.4295 & 0.4765 
			\\
			
			\makecell[c]{PspNet} 	
			& 0.0182 & 0.8810 & 0.0357 
			& 0.0775 & 0.6564 & 0.7339 
			\\
			
			\makecell[c]{DeepLab} 	
			& 0.0169 & 0.8751 & 0.0331 
			& 0.0742 & 0.5394 & 0.6136 
			\\
			
			\makecell[c]{U2Net} 	
			& 0.0677 & 0.7685 & 0.1244 
			& 0.1289 & 0.3774 & 0.5063
			\\
			
			\makecell[c]{EANet} 	
			& 0.0024 & 0.8264 & 0.0048 
			& 0.1061 & 0.5206 & 0.6267
			\\
			
			\makecell[c]{ABCBet} 	
			& 0.0350 & 0.8032 & 0.0671 
			& 0.1187 & 0.4549 & 0.5736
			\\
			
			\makecell[c]{MAResUNet} 	
			& 0.0288 & 0.7858 & 0.0556 
			& 0.1271 & 0.3302 & 0.4573
			\\
			
			\makecell[c]{UNetFormer} 	
			& 0.0019 & 0.7837 & 0.0038 
			& 0.1324 & 0.7611 & 0.8935
			\\
			
			\makecell[c]{CCNet} 	
			& 0.0060 & 0.8581 & 0.0120 
			& 0.0834 & 0.5953 & 0.6787
			\\
			
			\makecell[c]{SegFormer} 	
			& 0.0235 & 0.8200 & 0.0458
			& 0.1097 & 0.4691 & 0.5788
			\\

			\makecell[c]{UNet++} 	
			& 0.1307 & 0.7692 & 0.2234
			& 0.1295 & 0.1718 & 0.3013
			\\
			
			\makecell[c]{DenseAspp} 	
			& 0.0058 & 0.6871 & 0.0114
			& 0.1689 & 0.5051 & 0.6740
			\\
			
			\makecell[c]{ENet} 	
			& 0.0302 & 0.9278 & 0.0585
			& 0.0458 & 0.5962 & 0.6420
			\\
			
			\hline
			\makecell[c]{Ours (DeepMerge)} 
			& 0.9679 & \textbf{0.9425} & \textbf{0.9550} 
			& \textbf{0.0454} & 0.0441 & \textbf{0.0895} 
			\\
			\hline
		\end{tabular}
		
		\label{results}
	\end{table}
	
	Example segmentation results of the investigated algorithms are presented in Fig.\ref{resultsInResidence}, where three typical landscape types are selected: 1) urban residential areas (Fig.\ref{resultsInResidence}a), 2) rural industrial zones (Fig.\ref{resultsInIndustry}b), and 3) rural green spaces (Fig.\ref{resultsInRural}c). We notice that results from MRS contain notable over-segmentation errors. Relying on the results from MRS results, other investigated methods started to generate their optimized segmentations. We observe that the segmentation results of DeepMerge are satisfactory and superior to the others. Almost all house outlines, road boundaries, green spaces, and even sidewalk boundaries are precisely delineated by DeepMerge (Fig.\ref{resultsInResidence}a). FHS and USIH also achieve good performances, however, with discontinuous roads. The excellent performance of DeepMerge is also demonstrated in Fig.\ref{resultsInIndustry}b, where objects vary greatly in size, texture, shape, and spectrum. DeepMerge successfully delineates large areas of bare land, factory buildings, roads, and even individual trees. Local-SA and USIH also present satisfactory segmented results, but some over-segmentation errors still exist (see the spots in buildings of Local-SA and the bare lands of USIH). The above qualitative results well support the robustness of the proposed DeepMerge.\par
	
	\begin{landscape}
		
		\begin{figure}
			\ContinuedFloat*
			\centering\includegraphics[width=1.\linewidth]{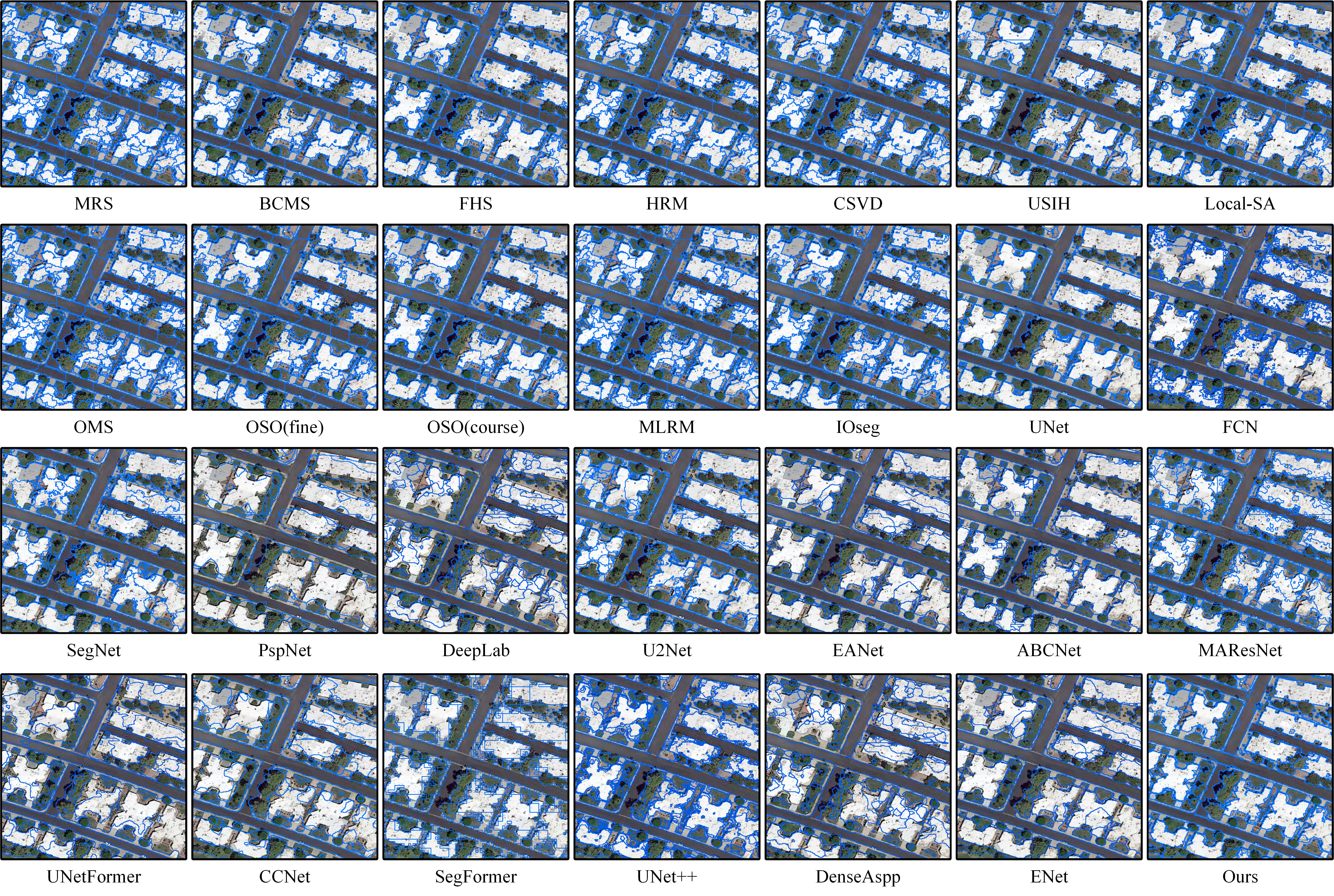}
			\caption{Image segmentation results in urban residential areas.}
			\label{resultsInResidence}
		\end{figure}

		
		\begin{figure}
			\ContinuedFloat
			\centering\includegraphics[width=1.\linewidth]{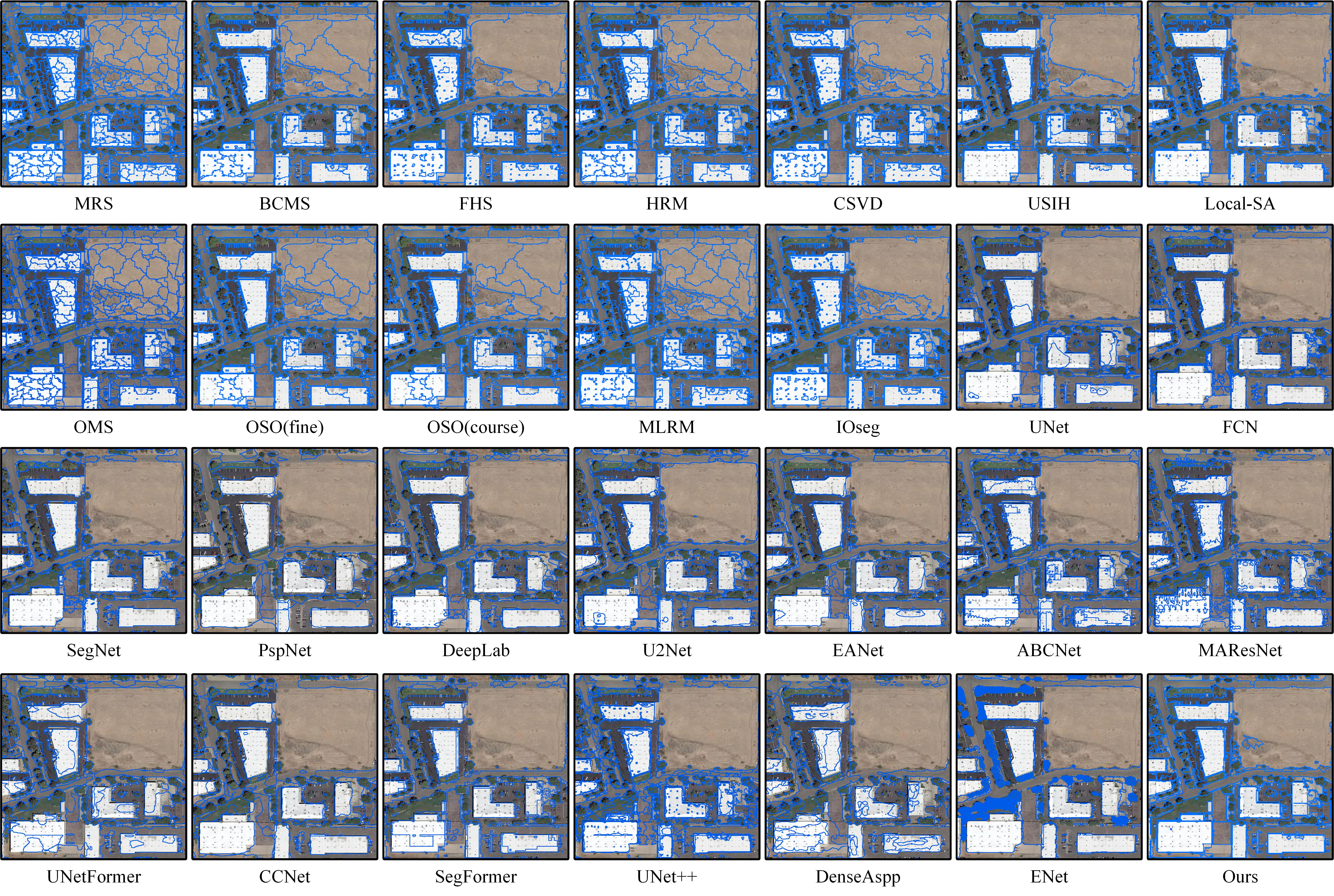}
			\caption{Image segmentation results in rural industrial zones.}
			\label{resultsInIndustry}
		\end{figure}
		\begin{figure}
			\ContinuedFloat
			\centering\includegraphics[width=1.\linewidth]{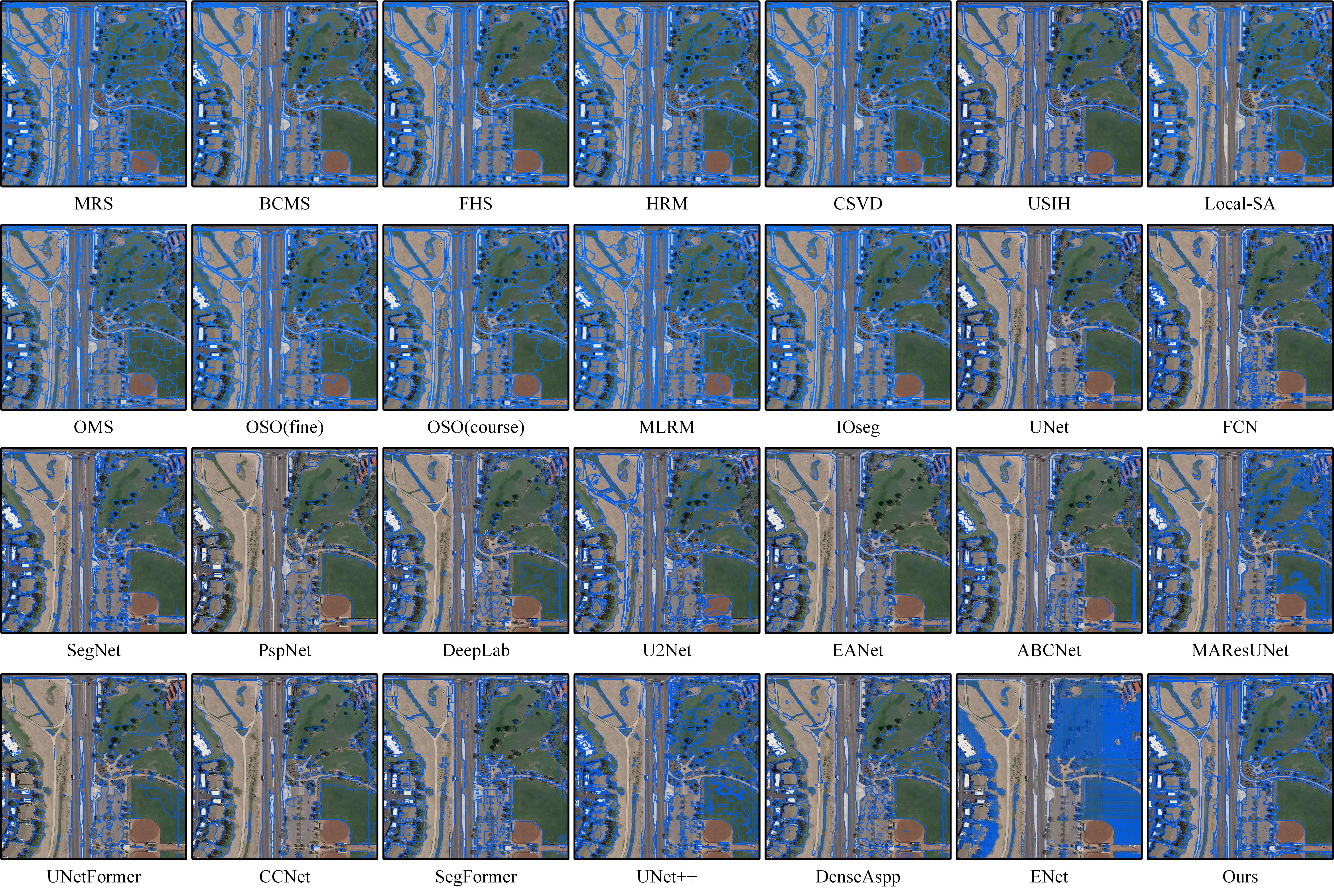}
			\caption{Image segmentation results in and rural green spaces.}
			\label{resultsInRural}
		\end{figure}
		
	\end{landscape}
	
	\subsection{Optimal scale parameters in DeepMerge}
	\label{scale-analysis}
\begin{figure}[h]
		\centering\includegraphics[width=1.0\linewidth]{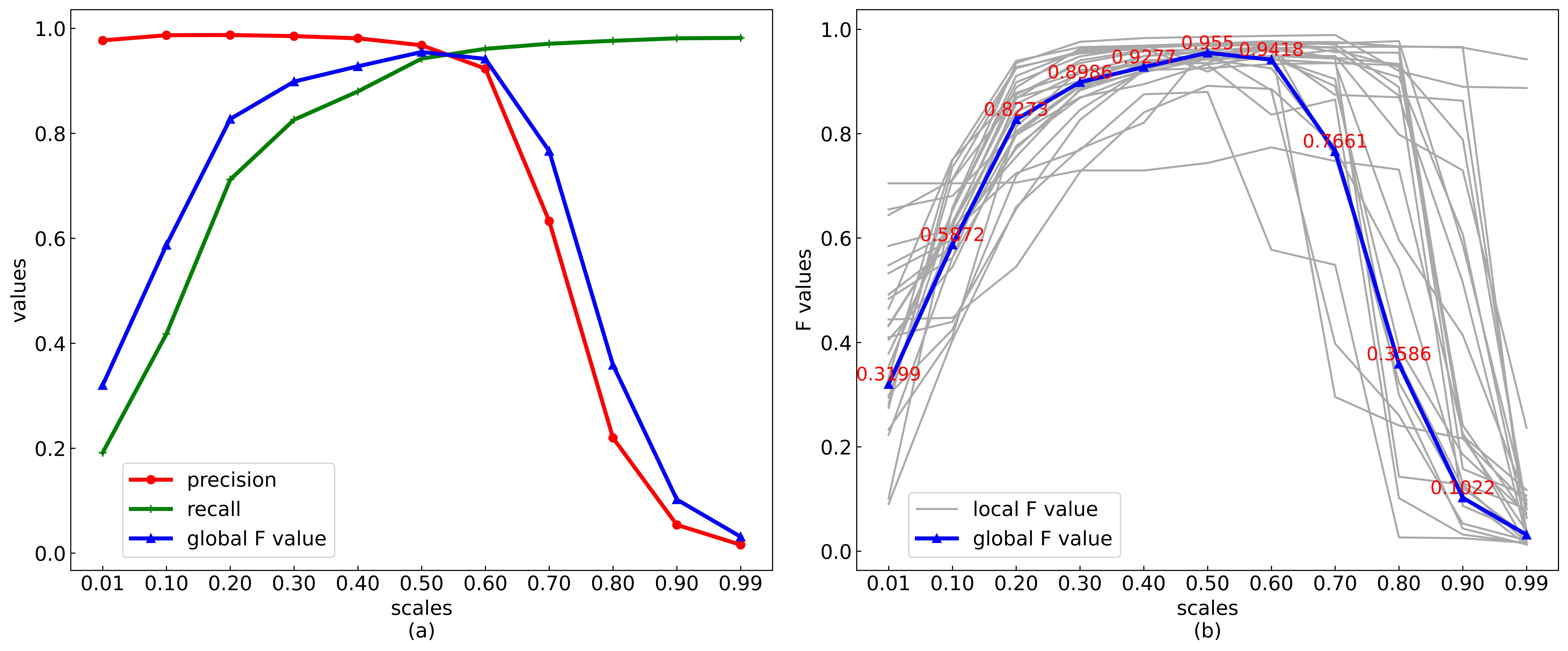}
		\caption{Segmentation performance of the proposed DeepMerge with different scale settings. (a) Performance curves by \emph{precision}, \emph{recall}, \emph{F}. (b) Performance curves by global and local F values in the study areas.}
		\label{resultsAtScales}
	\end{figure}

	The final segmentation results of the proposed method can be derived by setting different scale parameters. The segmentation evaluation metrics of \emph{precision}, \emph{recall}, and \emph{F}of the proposed DeepMerge are presented in Fig.\ref{resultsAtScales}a, where the scale parameters are set as 0.01, 0.10, 0.20, 0.30, 0.40, 0.50, 0.60, 0.70, 0.80, 0.90, and 0.99. The curve of \emph{precision} (red line in Fig.\ref{resultsAtScales}a) tends to decrease slowly for scale parameters from 0.0 to 0.6. After 0.6, however, the \emph{precision} decrease sharply with the further increase of scale parameters. The \emph{recall} value of the proposed DeepMerge increases sharply when the scale parameter is above 0.2. In comparison, the \emph{F} values first increase and then decrease with the continuous increase of scale parameters. The highest \emph{F} value is achieved when the scale parameter is 0.5. In general, the segmentation performance of DeepMerge varies with different settings of scale parameters. Fig.\ref{resultsAtScales}b depicts the global and local F values in the varying scale parameters, where the gray curves of local F values firstlt tend to increase and then decrease like global F values. All of the optimal scale parameters of local F values are distributed around 0.5. The above results suggest that the optimal scale for DeepMerge is 0.5 as Section\ref{loss-function} predicted.\par
	
	\subsection{Ablation experiments}
	
	The parameters involving margin value, training epoch number, batch size, and learning rate of DeepMerge are set to 1.0, 100, 120, and 0.0001, respectively. The quantitative segmentation measures via six evaluation metrics of ablation experiments are described in Table \ref{ablation}. `S2E` denotes a model with shift-scale embedding module and `3DP` means 3D position embedding module. As expected, the S2E model achieves satisfactory performance as it can learn shift scale features of objects. `Aux` denotes a model with an auxiliary module added to the backbone. `SEF` denotes a model with a segment-based feature module added to the backbone. The 'depth' depicts the layer number of stages in S2Former. The results suggest desirable performance, evidenced by the high \emph{recall} and \emph{F} values and low \emph{GOSE}, \emph{TE} values, proving the ability of S2Former to improve the image segmentation performance.\par
	
	\begin{table}[h]\scriptsize 
		\centering
		\caption{Ablation experiments on model variations.}
		\resizebox{\linewidth}{!}{
			\begin{tabular}{l l l l l l l l l l l}
				\hline
				\textbf{\makecell[c]{S2E}}
				& \textbf{\makecell[c]{3DP}}
				& \textbf{\makecell[c]{Aux}}
				& \textbf{\makecell[c]{SEF}}
				& \textbf{\makecell[c]{depth}}
				& \textbf{\makecell[c]{precision$\uparrow$}}
				& \textbf{\makecell[c]{recall$\uparrow$}} 
				& \textbf{\makecell[c]{F$\uparrow$}}
				& \textbf{\makecell[c]{GOSE$\downarrow$}}
				& \textbf{\makecell[c]{GUSE$\downarrow$}}
				& \textbf{\makecell[c]{TE$\downarrow$}}\\
				\hline
				
				\checkmark & & & & [6,4,2]
				& 0.9142 & 0.9122 & 0.9132
				& 0.0534 & 0.0539 & 0.1073 \\
				
				\checkmark &\checkmark  & & & [6,4,2]
				& 0.9254 & 0.9370 & 0.9311
				& 0.0453 & 0.0594 & 0.1048 \\
				
				\checkmark & & \checkmark &  & [6,4,2]
				&0.9386 & 0.9056 & 0.9218
				& 0.056 & 0.0532 &0.1093 \\
				
				\checkmark & & & \checkmark & [6,4,2]
				&0.9464  &0.9178  &0.9319
				&0.0547 &0.0485  &0.1032  \\

				\checkmark & & \checkmark & \checkmark & [6,4,2]
				& 0.9483 & 0.9320 &0.9401
				& 0.048 & 0.0533 & 0.1014 \\
				
				\checkmark & \checkmark &  & \checkmark & [6,4,2]
				& 0.9483 & 0.9209 & 0.9344
				& 0.0543 & 0.0487 & 0.1031 \\
				
				\checkmark & \checkmark &\checkmark & &[6,4,2]
				& 0.9246 & \textbf{0.9431} & 0.9338 
				& \textbf{0.0415} &  0.0551 & 0.0966 \\
				
				\checkmark & \checkmark & \checkmark &\checkmark &[6,4,2]
				& \textbf{0.9679} & 0.9425 & \textbf{0.9550} 
				& 0.0454 & \textbf{0.0441} & \textbf{0.0895} \\
				
				\checkmark & \checkmark & \checkmark &\checkmark &[3,2,1]
				& 0.8823 & 0.9077 & 0.8948 
				& 0.0580 & 0.0639 & 0.1219 \\
				\hline
			\end{tabular}
		}
		
		\label{ablation}
	\end{table}
	\section{Discussion}
	\label{S:5}
	
	The proposed DeepMerge only takes 0.19\% of the total number of super-pixels as training to achieve desirable image segmentation results. The optimal scale parameter value stabilizes at 0.5 in the study areas, releasing the selection of the optimal scale parameter value for users. Thus, DeepMerge overcomes scale parameter selection, unlike other multi-scale segmentation methods. A closer-to-zero scale parameter denotes a high similarity, while values close to or greater than one denote a low similarity. From Fig.\ref{resultsAtScales}, we notice that segmentation results in the scale range of [0.4, 0.6] are desirable for many applications, and the proposed DeepMerge can generate segmentation results in different scale parameters as needed.\par
	
\begin{figure}[h]
		\centering\includegraphics[width=0.85\linewidth]{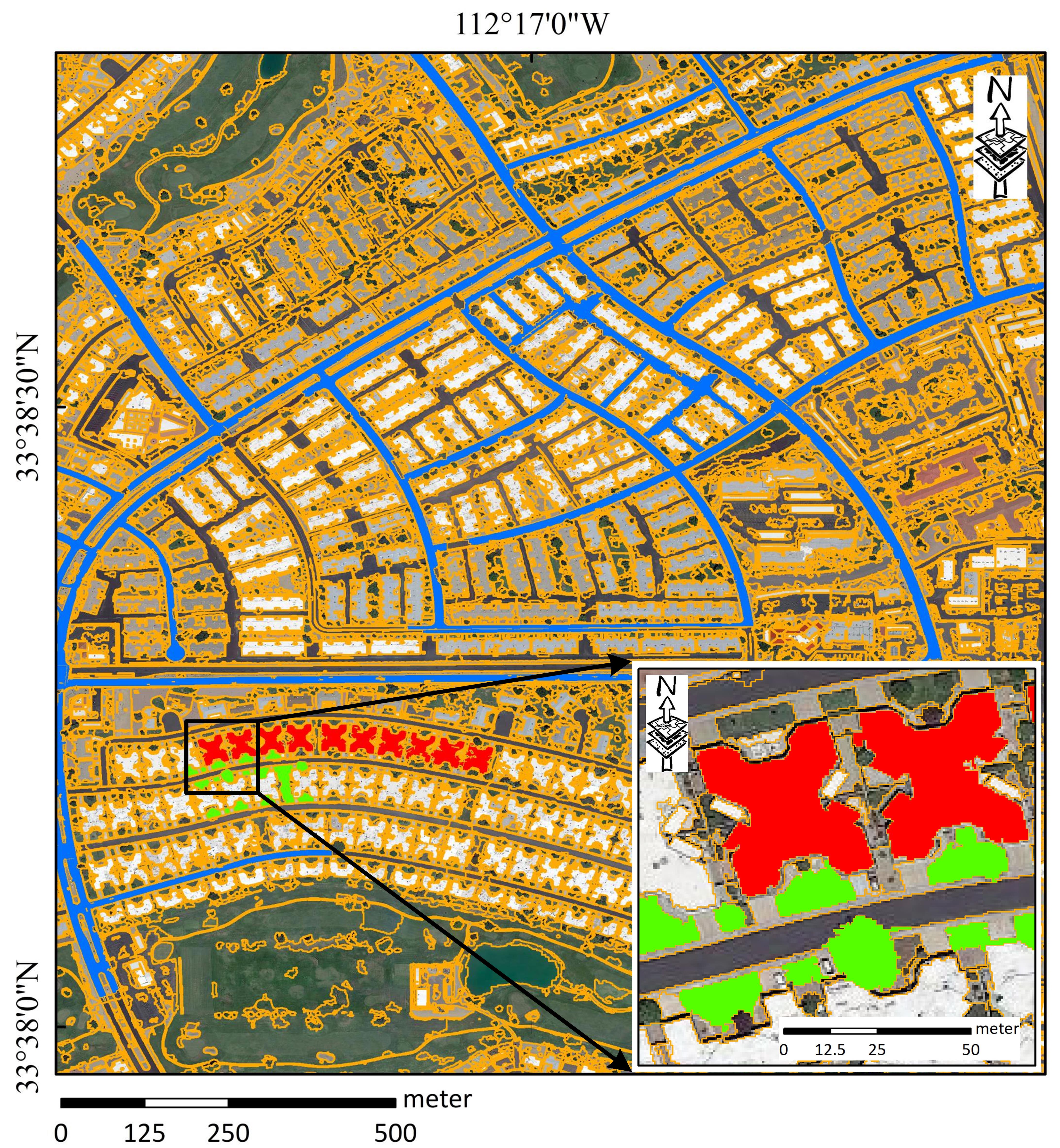}
		\caption{A case of objects with various sizes for presentation.}
		\label{big-small}
	\end{figure}

	To ensure the efficiency of the merging process, many region-merging methods introduce the object size as a parameter in the merging criteria. Although such a strategy improves the merging efficiency and generates even-sized objects, it fails to meet the demand of many applications due to the uneven size distribution of objects in an image. For example, the tiny objects in our dataset include individual trees and three-pixel-wide intermittent sidewalks. In comparison, factory buildings and roads usually contain thousands of pixels as shown in Fig.\ref{big-small}. The blue mask in the figure is the road segment as a whole object with 166,981m$^{2}$ in size. The polygons in red are independent residential houses with an average area of 1,000m$^{2}$. In the green areas, there are individual lawns, in which the smallest area of lawns only covers 42m$^{2}$. In this case, the blue road area is 3,976 times the size of the smaller lawn. In this study, the segmentation results of DeepMerge are able to capture the true sizes of objects.\par
	
	\section{Conclusion}
	\label{S:6}
	
	In this study, we proposed a deep step-wise optimization method for image segmentation of large-area and high-resolution remote sensing imagery. Given an initial super-pixel segmentation result, DeepMerge automatically obtains optimal segmentation output as a vector format in an interpretable scale parameter.\par
	
	Our method combines deep learning and RAG. We proposed the shift-scale embedding, shift-scale attention mechanism, and interpretable scale parameter to form S2Former to capture the shift-scale information. In addition, we introduced the segment-based feature embedding module into the networks to hold the object features. The segmentation performance on the Phoenix city cluster of our proposed method performs out SOTA methods in both qualitative and quantitative measurements. The proposed method requires a small number of training samples with respect to the total number of super-pixels. In our experimental analysis, only 0.19\% of the total number of super-pixel were labelled. DeepMerge based on a low ratio training dataset, achieves high advancements of 0.1 in \emph{F} value, large decrease of 0.04 in \emph{TE} against SOTA methods. The optimal scale parameter of the proposed method stabilizes at 0.5. This makes the selection of the optimal parameter value easier for the user. The proposed DeepMerge is suitable for the precise segmentation of large-area and very high-spatial-resolution remote sensing images. It will provide efficient and precise segmentation of products. However, it still requires training samples, leading to time-consuming segmentation in small areas.\par
	
	We plan to direct our further works to the potential improvement and the wide applications of the proposed DeepMerge by exploring the possibility of developing unsupervised DeepMerge-based segmentation approaches and further evaluating DeepMerge’s performance on various land-cover classification problems.
	
	\section{Acknowledgements}
	\label{S:7}
	This work was supported in part by the China Scholarship Council, Scientific Research Startup Fund of Northeastern University at Qinhuangdao.
	
	\appendix
	\section{Detials about semantic segmentation}
	\label{details-for-e2e}

\begin{figure}[h]
		\centering\includegraphics[width=0.69\linewidth]{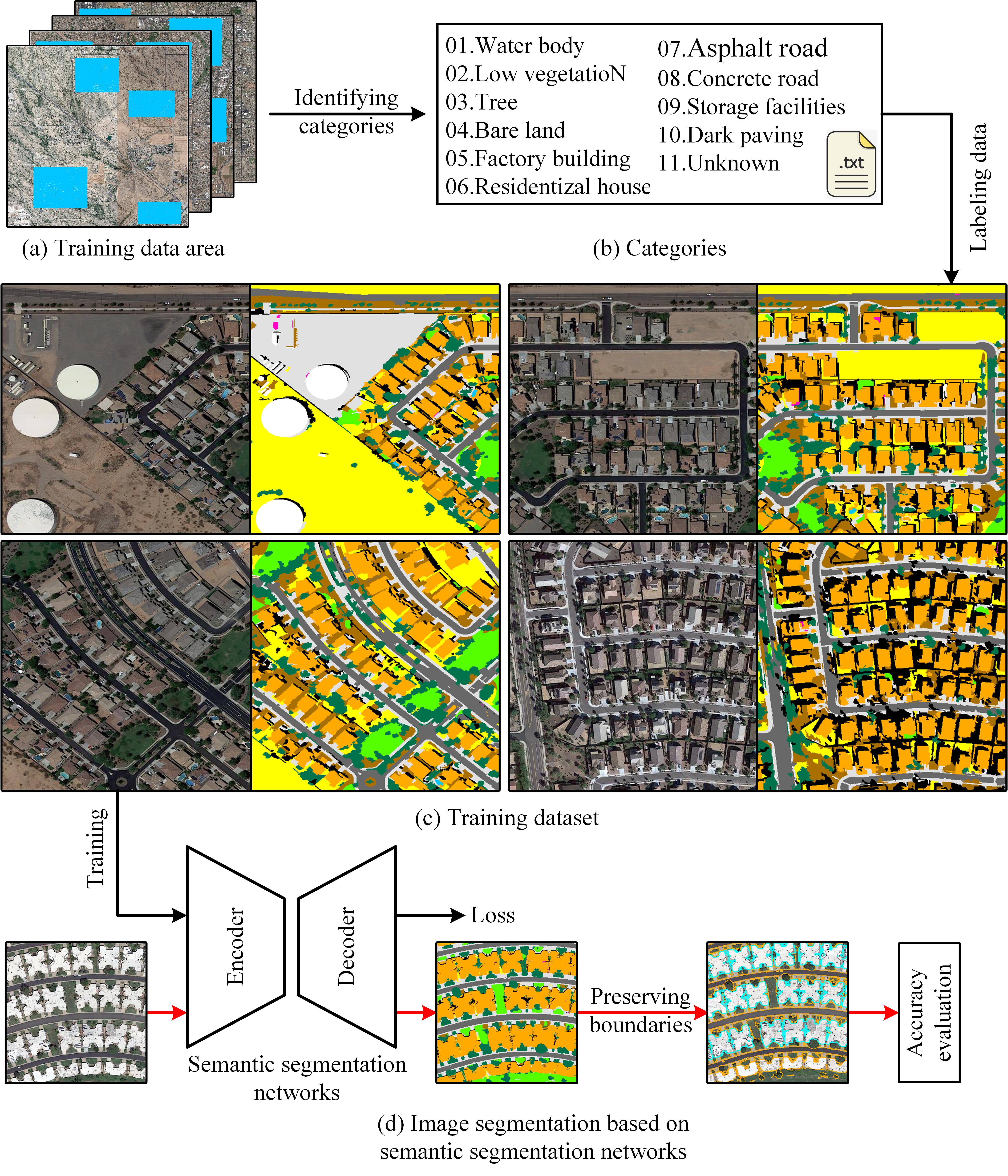}
		\caption{Workflow aboout image segmentatin based on semantic segmentation networks. (a) is the training data area corresponding to Fig.\ref{workflow}; (b) shows the land cover categories in the dataset; (c) is the training datset for semantic segmentation; (d) is the processing of remote sensing image segmentation based on semantic segmentaiton methods.}
		\label{dataset-for-e2e}
	\end{figure}

	Fig.\ref{dataset-for-e2e} depicts the workflow of labeling dataset, training semantic segmentation, preserving boundaries, and segmentation accuracy estimation for semantic segmentation networks. To ensure the consistency of training data, all super-pixels in the blue tiles are selected as the training dataset for semantic segmentation methods. We manually label the data under the blue zones as the training dataset, containing 11 classes (shown in Fig.\ref{dataset-for-e2e}b) and 1,803 training images with 512$\times$512 in pixel size.  After training the semantic segmentation networks, the output in the tested is the prediction result of pixel-wise category. To make the results comparable, we only preserved the boundaries of the results and saved it as standard segmentation outputs shown in Fig.\ref{dataset-for-e2e}d.\par
\begin{figure}[h]
		\centering\includegraphics[width=0.69\linewidth]{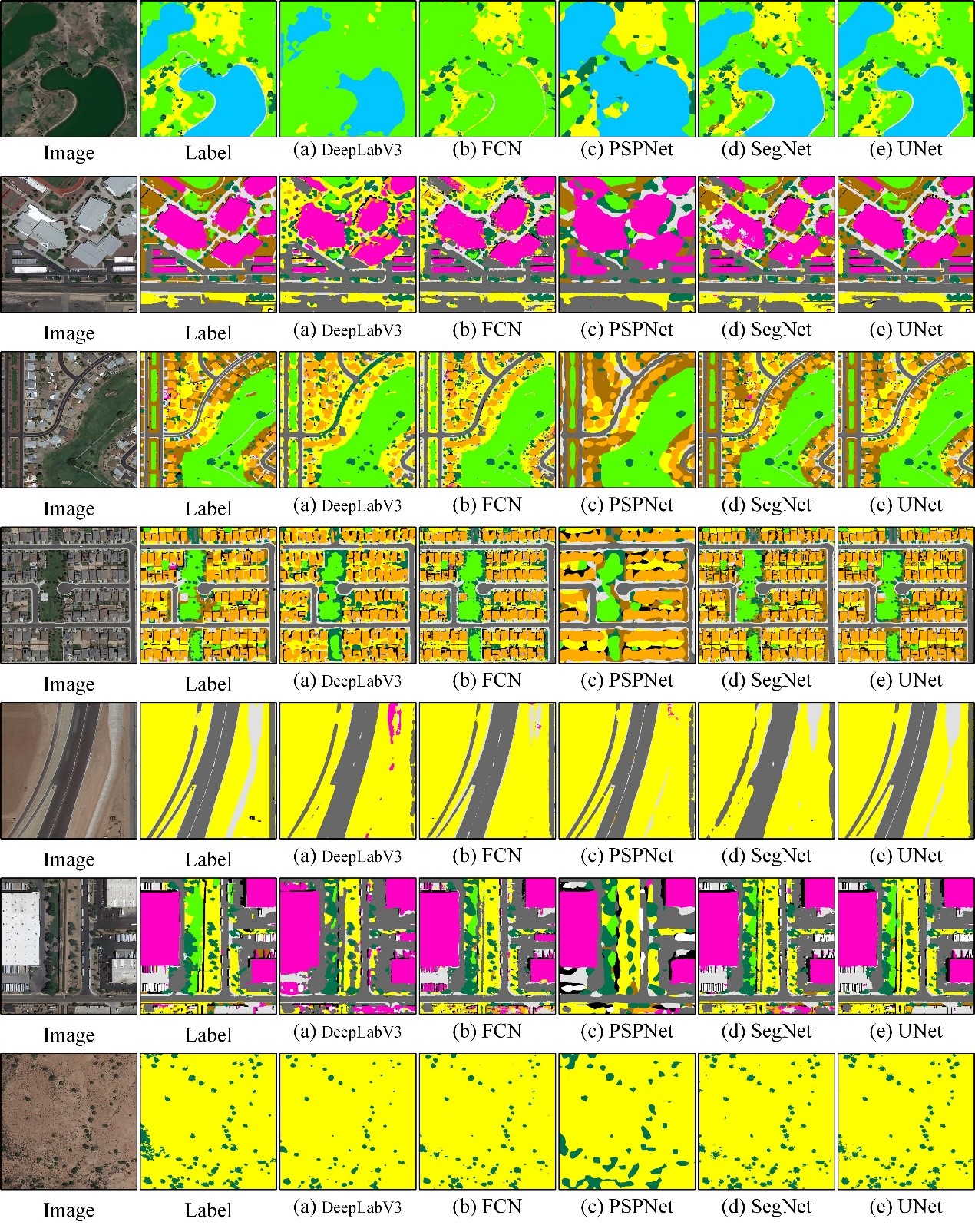}
		\caption{Semantic segmentation results of five standard networks.}
		\label{exampels-for-e2eresults}
	\end{figure}

	Based on the dataset, the results of semantic segmentation methods are not bad if F score based on pixels was used in the test area shown in Fig.\ref{exampels-for-e2eresults}.  For example, the F score of UNet is high in 0.7085 in the test area. However, the images segmentation accessment metric is F values based reference polygons different from F scores based on pixels. The precision and recall measures are calculated based on region overlapping. The matching direction for the precision measure is defined as a reference-to-segment directional correspondence. For the recall measure, we reverse this and match segments to reference objects.\par
\begin{figure}[h]
		\centering\includegraphics[width=1.0\linewidth]{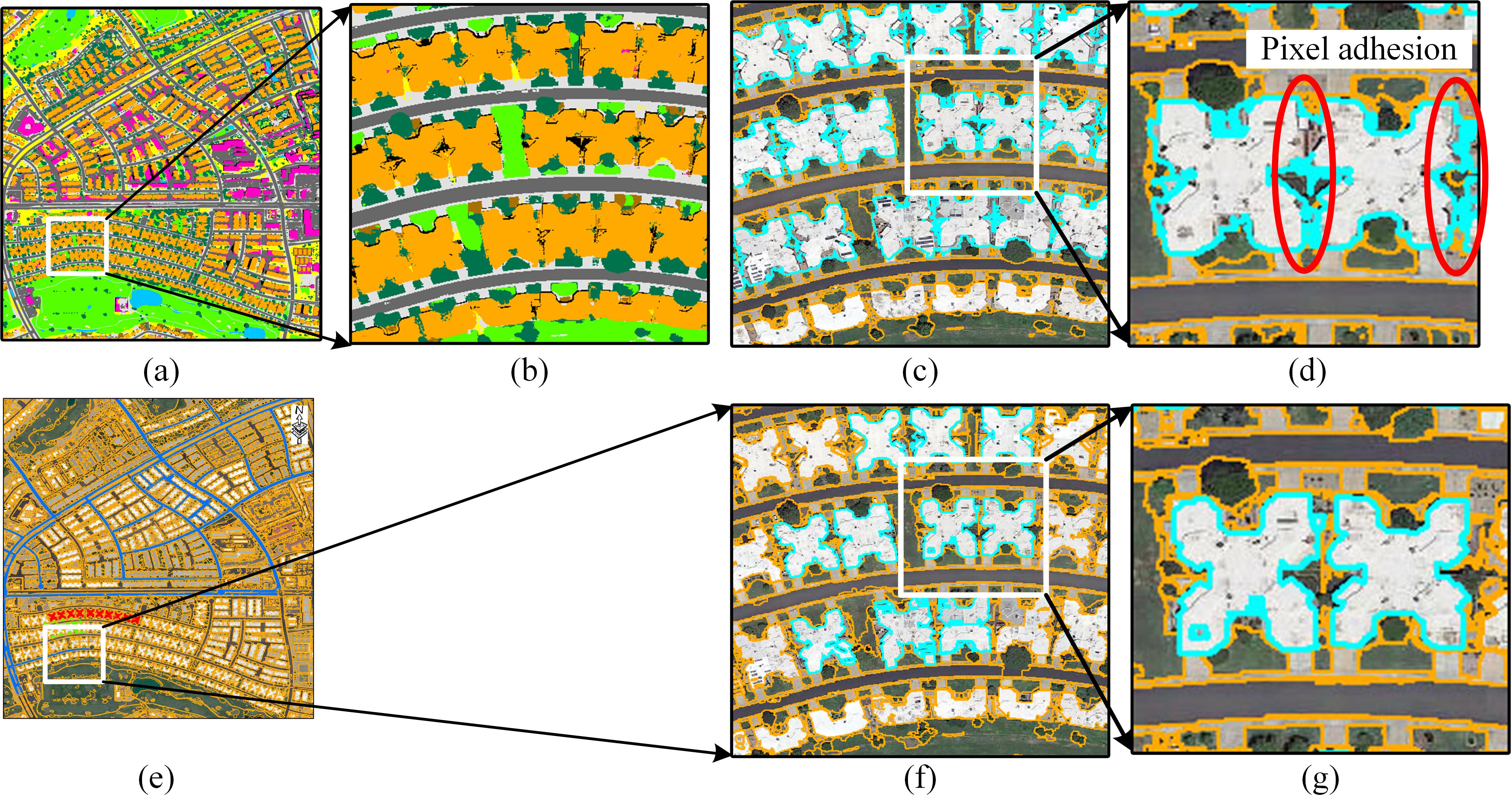}
		\caption{Pixel adhension of semantic segmentation methods.(a) and (e) are the output of semantic segmentation and image segmentation; (b) is a zoom in diagram with a local region; (c) and (f) are image segmentation results; (d) is the image segmentation result of semantic segmentation showing pixel adhension; (g) is the image segmentation of our method.}
		\label{pixel-adhesion}
	\end{figure}

	Pixel adhension often occurs in similar neighbouring objects in the semantic segmentation results shown in Fig.\ref{pixel-adhesion}. Though the semantic segmentation in Fig.\ref{pixel-adhesion}b is desirable by visual accessment, the image segmentation results based on their boundaries are bad, causing the low F values in the Table.\ref{results}.\par
	\clearpage
	
	
	
	
	
	
	\bibliographystyle{elsarticle-num-names}
	\bibliography{reference.bib}

\begin{thebibliography}{61}
\expandafter\ifx\csname natexlab\endcsname\relax\def\natexlab#1{#1}\fi
\providecommand{\url}[1]{\texttt{#1}}
\providecommand{\href}[2]{#2}
\providecommand{\path}[1]{#1}
\providecommand{\DOIprefix}{doi:}
\providecommand{\ArXivprefix}{arXiv:}
\providecommand{\URLprefix}{URL: }
\providecommand{\Pubmedprefix}{pmid:}
\providecommand{\doi}[1]{\href{http://dx.doi.org/#1}{\path{#1}}}
\providecommand{\Pubmed}[1]{\href{pmid:#1}{\path{#1}}}
\providecommand{\bibinfo}[2]{#2}
\ifx\xfnm\relax \def\xfnm[#1]{\unskip,\space#1}\fi
\bibitem[{Lv et~al.(2021)Lv, Shao, Ming, Diao, Zhou, and Tong}]{lv2021improved}
\bibinfo{author}{X.~Lv}, \bibinfo{author}{Z.~Shao}, \bibinfo{author}{D.~Ming},
  \bibinfo{author}{C.~Diao}, \bibinfo{author}{K.~Zhou},
  \bibinfo{author}{C.~Tong},
\newblock \bibinfo{title}{Improved object-based convolutional neural network
  (iocnn) to classify very high-resolution remote sensing images},
\newblock \bibinfo{journal}{International Journal of Remote Sensing}
  \bibinfo{volume}{42} (\bibinfo{year}{2021}) \bibinfo{pages}{8318--8344}.
\bibitem[{Zhou et~al.(2020)Zhou, Ming, Lv, Zhou, Bao, and Hong}]{zhou2020so}
\bibinfo{author}{W.~Zhou}, \bibinfo{author}{D.~Ming}, \bibinfo{author}{X.~Lv},
  \bibinfo{author}{K.~Zhou}, \bibinfo{author}{H.~Bao},
  \bibinfo{author}{Z.~Hong},
\newblock \bibinfo{title}{So--cnn based urban functional zone fine division
  with vhr remote sensing image},
\newblock \bibinfo{journal}{Remote Sensing of Environment}
  \bibinfo{volume}{236} (\bibinfo{year}{2020}) \bibinfo{pages}{111458}.
\bibitem[{Mundia and Aniya(2005)}]{mundia2005analysis}
\bibinfo{author}{C.~N. Mundia}, \bibinfo{author}{M.~Aniya},
\newblock \bibinfo{title}{Analysis of land use/cover changes and urban
  expansion of nairobi city using remote sensing and gis},
\newblock \bibinfo{journal}{International journal of Remote sensing}
  \bibinfo{volume}{26} (\bibinfo{year}{2005}) \bibinfo{pages}{2831--2849}.
\bibitem[{Zhao et~al.(2021)Zhao, Persello, and Stein}]{zhao2021building}
\bibinfo{author}{W.~Zhao}, \bibinfo{author}{C.~Persello},
  \bibinfo{author}{A.~Stein},
\newblock \bibinfo{title}{Building outline delineation: From aerial images to
  polygons with an improved end-to-end learning framework},
\newblock \bibinfo{journal}{ISPRS journal of photogrammetry and remote sensing}
  \bibinfo{volume}{175} (\bibinfo{year}{2021}) \bibinfo{pages}{119--131}.
\bibitem[{Na et~al.(2021)Na, Ding, Zhao, Liu, Tang, and Pfeifer}]{na2021object}
\bibinfo{author}{J.~Na}, \bibinfo{author}{H.~Ding}, \bibinfo{author}{W.~Zhao},
  \bibinfo{author}{K.~Liu}, \bibinfo{author}{G.~Tang},
  \bibinfo{author}{N.~Pfeifer},
\newblock \bibinfo{title}{Object-based large-scale terrain classification
  combined with segmentation optimization and terrain features: A case study in
  china},
\newblock \bibinfo{journal}{Transactions in GIS} \bibinfo{volume}{25}
  (\bibinfo{year}{2021}) \bibinfo{pages}{2939--2962}.
\bibitem[{Chen et~al.(2021)Chen, Zhong, Zheng, Ma, and Lu}]{chen2021urban}
\bibinfo{author}{D.~Chen}, \bibinfo{author}{Y.~Zhong},
  \bibinfo{author}{Z.~Zheng}, \bibinfo{author}{A.~Ma}, \bibinfo{author}{X.~Lu},
\newblock \bibinfo{title}{Urban road mapping based on an end-to-end road
  vectorization mapping network framework},
\newblock \bibinfo{journal}{ISPRS Journal of Photogrammetry and Remote Sensing}
  \bibinfo{volume}{178} (\bibinfo{year}{2021}) \bibinfo{pages}{345--365}.
\bibitem[{Zhao et~al.(2022)Zhao, Persello, and Stein}]{zhao2022extracting}
\bibinfo{author}{W.~Zhao}, \bibinfo{author}{C.~Persello},
  \bibinfo{author}{A.~Stein},
\newblock \bibinfo{title}{Extracting planar roof structures from very high
  resolution images using graph neural networks},
\newblock \bibinfo{journal}{ISPRS Journal of Photogrammetry and Remote Sensing}
  \bibinfo{volume}{187} (\bibinfo{year}{2022}) \bibinfo{pages}{34--45}.
\bibitem[{Blaschke(2010)}]{blaschke2010object}
\bibinfo{author}{T.~Blaschke},
\newblock \bibinfo{title}{Object based image analysis for remote sensing},
\newblock \bibinfo{journal}{ISPRS journal of photogrammetry and remote sensing}
  \bibinfo{volume}{65} (\bibinfo{year}{2010}) \bibinfo{pages}{2--16}.
\bibitem[{Zhang et~al.(2014)Zhang, Xiao, and Feng}]{zhang2014fast}
\bibinfo{author}{X.~Zhang}, \bibinfo{author}{P.~Xiao},
  \bibinfo{author}{X.~Feng},
\newblock \bibinfo{title}{Fast hierarchical segmentation of high-resolution
  remote sensing image with adaptive edge penalty},
\newblock \bibinfo{journal}{Photogrammetric Engineering \& Remote Sensing}
  \bibinfo{volume}{80} (\bibinfo{year}{2014}) \bibinfo{pages}{71--80}.
\bibitem[{Beaulieu and Goldberg(1989)}]{beaulieu1989hierarchy}
\bibinfo{author}{J.-M. Beaulieu}, \bibinfo{author}{M.~Goldberg},
\newblock \bibinfo{title}{Hierarchy in picture segmentation: A stepwise
  optimization approach},
\newblock \bibinfo{journal}{IEEE Transactions on pattern analysis and machine
  intelligence} \bibinfo{volume}{11} (\bibinfo{year}{1989})
  \bibinfo{pages}{150--163}.
\bibitem[{Haris et~al.(1998)Haris, Efstratiadis, Maglaveras, and
  Katsaggelos}]{haris1998hybrid}
\bibinfo{author}{K.~Haris}, \bibinfo{author}{S.~N. Efstratiadis},
  \bibinfo{author}{N.~Maglaveras}, \bibinfo{author}{A.~K. Katsaggelos},
\newblock \bibinfo{title}{Hybrid image segmentation using watersheds and fast
  region merging},
\newblock \bibinfo{journal}{IEEE Transactions on image processing}
  \bibinfo{volume}{7} (\bibinfo{year}{1998}) \bibinfo{pages}{1684--1699}.
\bibitem[{Zhang et~al.(2013)Zhang, Xiao, Song, and She}]{zhang2013boundary}
\bibinfo{author}{X.~Zhang}, \bibinfo{author}{P.~Xiao},
  \bibinfo{author}{X.~Song}, \bibinfo{author}{J.~She},
\newblock \bibinfo{title}{Boundary-constrained multi-scale segmentation method
  for remote sensing images},
\newblock \bibinfo{journal}{ISPRS Journal of Photogrammetry and Remote Sensing}
  \bibinfo{volume}{78} (\bibinfo{year}{2013}) \bibinfo{pages}{15--25}.
\bibitem[{Yang et~al.(2017)Yang, He, and Caspersen}]{yang2017region}
\bibinfo{author}{J.~Yang}, \bibinfo{author}{Y.~He},
  \bibinfo{author}{J.~Caspersen},
\newblock \bibinfo{title}{Region merging using local spectral angle thresholds:
  A more accurate method for hybrid segmentation of remote sensing images},
\newblock \bibinfo{journal}{Remote sensing of environment}
  \bibinfo{volume}{190} (\bibinfo{year}{2017}) \bibinfo{pages}{137--148}.
\bibitem[{Achanta et~al.(2012)Achanta, Shaji, Smith, Lucchi, Fua, and
  S{\"u}sstrunk}]{achanta2012slic}
\bibinfo{author}{R.~Achanta}, \bibinfo{author}{A.~Shaji},
  \bibinfo{author}{K.~Smith}, \bibinfo{author}{A.~Lucchi},
  \bibinfo{author}{P.~Fua}, \bibinfo{author}{S.~S{\"u}sstrunk},
\newblock \bibinfo{title}{Slic superpixels compared to state-of-the-art
  superpixel methods},
\newblock \bibinfo{journal}{IEEE transactions on pattern analysis and machine
  intelligence} \bibinfo{volume}{34} (\bibinfo{year}{2012})
  \bibinfo{pages}{2274--2282}.
\bibitem[{Paris and Durand(2007)}]{paris2007topological}
\bibinfo{author}{S.~Paris}, \bibinfo{author}{F.~Durand},
\newblock \bibinfo{title}{A topological approach to hierarchical segmentation
  using mean shift},
\newblock in: \bibinfo{booktitle}{2007 IEEE Conference on Computer Vision and
  Pattern Recognition}, \bibinfo{organization}{IEEE}, \bibinfo{year}{2007}, pp.
  \bibinfo{pages}{1--8}.
\bibitem[{Baatz(2000)}]{baatz2000multi}
\bibinfo{author}{M.~Baatz},
\newblock \bibinfo{title}{Multi resolution segmentation: an optimum approach
  for high quality multi scale image segmentation},
\newblock in: \bibinfo{booktitle}{Beutrage zum AGIT-Symposium. Salzburg,
  Heidelberg, 2000}, \bibinfo{year}{2000}, pp. \bibinfo{pages}{12--23}.
\bibitem[{Martin et~al.(2004)Martin, Fowlkes, and Malik}]{martin2004learning}
\bibinfo{author}{D.~R. Martin}, \bibinfo{author}{C.~C. Fowlkes},
  \bibinfo{author}{J.~Malik},
\newblock \bibinfo{title}{Learning to detect natural image boundaries using
  local brightness, color, and texture cues},
\newblock \bibinfo{journal}{IEEE transactions on pattern analysis and machine
  intelligence} \bibinfo{volume}{26} (\bibinfo{year}{2004})
  \bibinfo{pages}{530--549}.
\bibitem[{Arbelaez et~al.(2010)Arbelaez, Maire, Fowlkes, and
  Malik}]{arbelaez2010contour}
\bibinfo{author}{P.~Arbelaez}, \bibinfo{author}{M.~Maire},
  \bibinfo{author}{C.~Fowlkes}, \bibinfo{author}{J.~Malik},
\newblock \bibinfo{title}{Contour detection and hierarchical image
  segmentation},
\newblock \bibinfo{journal}{IEEE transactions on pattern analysis and machine
  intelligence} \bibinfo{volume}{33} (\bibinfo{year}{2010})
  \bibinfo{pages}{898--916}.
\bibitem[{Pont-Tuset et~al.(2016)Pont-Tuset, Arbelaez, Barron, Marques, and
  Malik}]{pont2016multiscale}
\bibinfo{author}{J.~Pont-Tuset}, \bibinfo{author}{P.~Arbelaez},
  \bibinfo{author}{J.~T. Barron}, \bibinfo{author}{F.~Marques},
  \bibinfo{author}{J.~Malik},
\newblock \bibinfo{title}{Multiscale combinatorial grouping for image
  segmentation and object proposal generation},
\newblock \bibinfo{journal}{IEEE transactions on pattern analysis and machine
  intelligence} \bibinfo{volume}{39} (\bibinfo{year}{2016})
  \bibinfo{pages}{128--140}.
\bibitem[{Derivaux et~al.(2006)Derivaux, Lefevre, Wemmert, and
  Korczak}]{derivaux2006watershed}
\bibinfo{author}{S.~Derivaux}, \bibinfo{author}{S.~Lefevre},
  \bibinfo{author}{C.~Wemmert}, \bibinfo{author}{J.~Korczak},
\newblock \bibinfo{title}{Watershed segmentation of remotely sensed images
  based on a supervised fuzzy pixel classification},
\newblock in: \bibinfo{booktitle}{2006 IEEE International Symposium on
  Geoscience and Remote Sensing}, \bibinfo{organization}{IEEE},
  \bibinfo{year}{2006}, pp. \bibinfo{pages}{3712--3715}.
\bibitem[{Wassenberg et~al.(2009)Wassenberg, Middelmann, and
  Sanders}]{wassenberg2009efficient}
\bibinfo{author}{J.~Wassenberg}, \bibinfo{author}{W.~Middelmann},
  \bibinfo{author}{P.~Sanders},
\newblock \bibinfo{title}{An efficient parallel algorithm for graph-based image
  segmentation},
\newblock in: \bibinfo{booktitle}{International Conference on Computer Analysis
  of Images and Patterns}, \bibinfo{organization}{Springer},
  \bibinfo{year}{2009}, pp. \bibinfo{pages}{1003--1010}.
\bibitem[{Johnson and Xie(2011)}]{johnson2011unsupervised}
\bibinfo{author}{B.~Johnson}, \bibinfo{author}{Z.~Xie},
\newblock \bibinfo{title}{Unsupervised image segmentation evaluation and
  refinement using a multi-scale approach},
\newblock \bibinfo{journal}{ISPRS Journal of Photogrammetry and Remote Sensing}
  \bibinfo{volume}{66} (\bibinfo{year}{2011}) \bibinfo{pages}{473--483}.
\bibitem[{Lee and Cok(1991)}]{lee1991detecting}
\bibinfo{author}{H.-C. Lee}, \bibinfo{author}{D.~R. Cok},
\newblock \bibinfo{title}{Detecting boundaries in a vector field},
\newblock \bibinfo{journal}{IEEE Transactions on Signal Processing}
  \bibinfo{volume}{39} (\bibinfo{year}{1991}) \bibinfo{pages}{1181--1194}.
\bibitem[{Su et~al.(2020)Su, Liu, Zhang, Qu, and Li}]{su2020machine}
\bibinfo{author}{T.~Su}, \bibinfo{author}{T.~Liu}, \bibinfo{author}{S.~Zhang},
  \bibinfo{author}{Z.~Qu}, \bibinfo{author}{R.~Li},
\newblock \bibinfo{title}{Machine learning-assisted region merging for remote
  sensing image segmentation},
\newblock \bibinfo{journal}{ISPRS Journal of Photogrammetry and Remote Sensing}
  \bibinfo{volume}{168} (\bibinfo{year}{2020}) \bibinfo{pages}{89--123}.
\bibitem[{Chen et~al.(2014)Chen, Deng, Mei, Chen, Shao, and
  Hong}]{chen2014optimal}
\bibinfo{author}{J.~Chen}, \bibinfo{author}{M.~Deng}, \bibinfo{author}{X.~Mei},
  \bibinfo{author}{T.~Chen}, \bibinfo{author}{Q.~Shao},
  \bibinfo{author}{L.~Hong},
\newblock \bibinfo{title}{Optimal segmentation of a high-resolution
  remote-sensing image guided by area and boundary},
\newblock \bibinfo{journal}{International Journal of Remote Sensing}
  \bibinfo{volume}{35} (\bibinfo{year}{2014}) \bibinfo{pages}{6914--6939}.
\bibitem[{Wang et~al.(2019)Wang, Qi, Liu, Jiang, and
  Wang}]{wang2019unsupervised}
\bibinfo{author}{Y.~Wang}, \bibinfo{author}{Q.~Qi}, \bibinfo{author}{Y.~Liu},
  \bibinfo{author}{L.~Jiang}, \bibinfo{author}{J.~Wang},
\newblock \bibinfo{title}{Unsupervised segmentation parameter selection using
  the local spatial statistics for remote sensing image segmentation},
\newblock \bibinfo{journal}{International Journal of Applied Earth Observation
  and Geoinformation} \bibinfo{volume}{81} (\bibinfo{year}{2019})
  \bibinfo{pages}{98--109}.
\bibitem[{Zheng et~al.(2020)Zheng, Du, Du, and Zhang}]{zheng2020multiscale}
\bibinfo{author}{Z.~Zheng}, \bibinfo{author}{S.~Du}, \bibinfo{author}{S.~Du},
  \bibinfo{author}{X.~Zhang},
\newblock \bibinfo{title}{A multiscale approach to delineate dune-field
  landscape patches},
\newblock \bibinfo{journal}{Remote Sensing of Environment}
  \bibinfo{volume}{237} (\bibinfo{year}{2020}) \bibinfo{pages}{111591}.
\bibitem[{Zhang et~al.(2018)Zhang, Sargent, Pan, Li, Gardiner, Hare, and
  Atkinson}]{zhang2018object}
\bibinfo{author}{C.~Zhang}, \bibinfo{author}{I.~Sargent},
  \bibinfo{author}{X.~Pan}, \bibinfo{author}{H.~Li},
  \bibinfo{author}{A.~Gardiner}, \bibinfo{author}{J.~Hare},
  \bibinfo{author}{P.~M. Atkinson},
\newblock \bibinfo{title}{An object-based convolutional neural network (ocnn)
  for urban land use classification},
\newblock \bibinfo{journal}{Remote sensing of environment}
  \bibinfo{volume}{216} (\bibinfo{year}{2018}) \bibinfo{pages}{57--70}.
\bibitem[{Dr{\u{a}}gu{\c{t}} et~al.(2014)Dr{\u{a}}gu{\c{t}}, Csillik, Eisank,
  and Tiede}]{druaguct2014automated}
\bibinfo{author}{L.~Dr{\u{a}}gu{\c{t}}}, \bibinfo{author}{O.~Csillik},
  \bibinfo{author}{C.~Eisank}, \bibinfo{author}{D.~Tiede},
\newblock \bibinfo{title}{Automated parameterisation for multi-scale image
  segmentation on multiple layers},
\newblock \bibinfo{journal}{ISPRS Journal of photogrammetry and Remote Sensing}
  \bibinfo{volume}{88} (\bibinfo{year}{2014}) \bibinfo{pages}{119--127}.
\bibitem[{Ming et~al.(2015)Ming, Li, Wang, and Zhang}]{ming2015scale}
\bibinfo{author}{D.~Ming}, \bibinfo{author}{J.~Li}, \bibinfo{author}{J.~Wang},
  \bibinfo{author}{M.~Zhang},
\newblock \bibinfo{title}{Scale parameter selection by spatial statistics for
  geobia: Using mean-shift based multi-scale segmentation as an example},
\newblock \bibinfo{journal}{ISPRS Journal of Photogrammetry and Remote Sensing}
  \bibinfo{volume}{106} (\bibinfo{year}{2015}) \bibinfo{pages}{28--41}.
\bibitem[{Hu et~al.(2018)Hu, Zhang, Zou, Li, and Wu}]{hu2018stepwise}
\bibinfo{author}{Z.~Hu}, \bibinfo{author}{Q.~Zhang}, \bibinfo{author}{Q.~Zou},
  \bibinfo{author}{Q.~Li}, \bibinfo{author}{G.~Wu},
\newblock \bibinfo{title}{Stepwise evolution analysis of the region-merging
  segmentation for scale parameterization},
\newblock \bibinfo{journal}{IEEE Journal of Selected Topics in Applied Earth
  Observations and Remote Sensing} \bibinfo{volume}{11} (\bibinfo{year}{2018})
  \bibinfo{pages}{2461--2472}.
\bibitem[{Zhang et~al.(2014)Zhang, Xiao, Feng, Wang, and
  Wang}]{zhang2014hybrid}
\bibinfo{author}{X.~Zhang}, \bibinfo{author}{P.~Xiao},
  \bibinfo{author}{X.~Feng}, \bibinfo{author}{J.~Wang},
  \bibinfo{author}{Z.~Wang},
\newblock \bibinfo{title}{Hybrid region merging method for segmentation of
  high-resolution remote sensing images},
\newblock \bibinfo{journal}{ISPRS Journal of Photogrammetry and Remote Sensing}
  \bibinfo{volume}{98} (\bibinfo{year}{2014}) \bibinfo{pages}{19--28}.
\bibitem[{Shen et~al.(2019)Shen, Chen, Xiao, and Pan}]{shen2019optimizing}
\bibinfo{author}{Y.~Shen}, \bibinfo{author}{J.~Chen},
  \bibinfo{author}{L.~Xiao}, \bibinfo{author}{D.~Pan},
\newblock \bibinfo{title}{Optimizing multiscale segmentation with local
  spectral heterogeneity measure for high resolution remote sensing images},
\newblock \bibinfo{journal}{ISPRS Journal of Photogrammetry and Remote Sensing}
  \bibinfo{volume}{157} (\bibinfo{year}{2019}) \bibinfo{pages}{13--25}.
\bibitem[{Zhang et~al.(2020)Zhang, Xiao, and Feng}]{zhang2020object}
\bibinfo{author}{X.~Zhang}, \bibinfo{author}{P.~Xiao},
  \bibinfo{author}{X.~Feng},
\newblock \bibinfo{title}{Object-specific optimization of hierarchical
  multiscale segmentations for high-spatial resolution remote sensing images},
\newblock \bibinfo{journal}{ISPRS Journal of Photogrammetry and Remote Sensing}
  \bibinfo{volume}{159} (\bibinfo{year}{2020}) \bibinfo{pages}{308--321}.
\bibitem[{Chopra et~al.(2005)Chopra, Hadsell, and LeCun}]{chopra2005learning}
\bibinfo{author}{S.~Chopra}, \bibinfo{author}{R.~Hadsell},
  \bibinfo{author}{Y.~LeCun},
\newblock \bibinfo{title}{Learning a similarity metric discriminatively, with
  application to face verification},
\newblock in: \bibinfo{booktitle}{2005 IEEE Computer Society Conference on
  Computer Vision and Pattern Recognition (CVPR'05)},
  volume~\bibinfo{volume}{1}, \bibinfo{organization}{IEEE},
  \bibinfo{year}{2005}, pp. \bibinfo{pages}{539--546}.
\bibitem[{Guo et~al.(2017)Guo, Feng, Zhou, Huang, Wan, and
  Wang}]{guo2017learning}
\bibinfo{author}{Q.~Guo}, \bibinfo{author}{W.~Feng}, \bibinfo{author}{C.~Zhou},
  \bibinfo{author}{R.~Huang}, \bibinfo{author}{L.~Wan},
  \bibinfo{author}{S.~Wang},
\newblock \bibinfo{title}{Learning dynamic siamese network for visual object
  tracking},
\newblock in: \bibinfo{booktitle}{Proceedings of the IEEE international
  conference on computer vision}, \bibinfo{year}{2017}, pp.
  \bibinfo{pages}{1763--1771}.
\bibitem[{Lv et~al.(2018)Lv, Ming, Lu, Zhou, Wang, and Bao}]{lv2018new}
\bibinfo{author}{X.~Lv}, \bibinfo{author}{D.~Ming}, \bibinfo{author}{T.~Lu},
  \bibinfo{author}{K.~Zhou}, \bibinfo{author}{M.~Wang},
  \bibinfo{author}{H.~Bao},
\newblock \bibinfo{title}{A new method for region-based majority voting cnns
  for very high resolution image classification},
\newblock \bibinfo{journal}{Remote Sensing} \bibinfo{volume}{10}
  (\bibinfo{year}{2018}) \bibinfo{pages}{1946}.
\bibitem[{Lv et~al.(2019)Lv, Ming, Chen, and Wang}]{lv2019very}
\bibinfo{author}{X.~Lv}, \bibinfo{author}{D.~Ming}, \bibinfo{author}{Y.~Chen},
  \bibinfo{author}{M.~Wang},
\newblock \bibinfo{title}{Very high resolution remote sensing image
  classification with seeds-cnn and scale effect analysis for superpixel cnn
  classification},
\newblock \bibinfo{journal}{International Journal of Remote Sensing}
  \bibinfo{volume}{40} (\bibinfo{year}{2019}) \bibinfo{pages}{506--531}.
\bibitem[{Lv et~al.(2022)Lv, Shao, Huang, Zhou, Ming, Wang, and
  Tong}]{lv2022bts}
\bibinfo{author}{X.~Lv}, \bibinfo{author}{Z.~Shao}, \bibinfo{author}{X.~Huang},
  \bibinfo{author}{W.~Zhou}, \bibinfo{author}{D.~Ming},
  \bibinfo{author}{J.~Wang}, \bibinfo{author}{C.~Tong},
\newblock \bibinfo{title}{Bts: a binary tree sampling strategy for object
  identification based on deep learning},
\newblock \bibinfo{journal}{International journal of geographical information
  science} \bibinfo{volume}{36} (\bibinfo{year}{2022})
  \bibinfo{pages}{822--848}.
\bibitem[{Arnab et~al.(2021)Arnab, Dehghani, Heigold, Sun, Lu{\v{c}}i{\'c}, and
  Schmid}]{arnab2021vivit}
\bibinfo{author}{A.~Arnab}, \bibinfo{author}{M.~Dehghani},
  \bibinfo{author}{G.~Heigold}, \bibinfo{author}{C.~Sun},
  \bibinfo{author}{M.~Lu{\v{c}}i{\'c}}, \bibinfo{author}{C.~Schmid},
\newblock \bibinfo{title}{Vivit: A video vision transformer},
\newblock in: \bibinfo{booktitle}{Proceedings of the IEEE/CVF International
  Conference on Computer Vision}, \bibinfo{year}{2021}, pp.
  \bibinfo{pages}{6836--6846}.
\bibitem[{Hendrycks and Gimpel(2016)}]{hendrycks2016gaussian}
\bibinfo{author}{D.~Hendrycks}, \bibinfo{author}{K.~Gimpel},
\newblock \bibinfo{title}{Gaussian error linear units (gelus)},
\newblock \bibinfo{journal}{arXiv preprint arXiv:1606.08415}
  (\bibinfo{year}{2016}).
\bibitem[{Zhang et~al.(2015)Zhang, Feng, Xiao, He, and
  Zhu}]{zhang2015segmentation}
\bibinfo{author}{X.~Zhang}, \bibinfo{author}{X.~Feng},
  \bibinfo{author}{P.~Xiao}, \bibinfo{author}{G.~He}, \bibinfo{author}{L.~Zhu},
\newblock \bibinfo{title}{Segmentation quality evaluation using region-based
  precision and recall measures for remote sensing images},
\newblock \bibinfo{journal}{ISPRS Journal of Photogrammetry and Remote Sensing}
  \bibinfo{volume}{102} (\bibinfo{year}{2015}) \bibinfo{pages}{73--84}.
\bibitem[{Su and Zhang(2017)}]{su2017local}
\bibinfo{author}{T.~Su}, \bibinfo{author}{S.~Zhang},
\newblock \bibinfo{title}{Local and global evaluation for remote sensing image
  segmentation},
\newblock \bibinfo{journal}{ISPRS Journal of Photogrammetry and Remote Sensing}
  \bibinfo{volume}{130} (\bibinfo{year}{2017}) \bibinfo{pages}{256--276}.
\bibitem[{Yu and Gong(2012)}]{yu2012google}
\bibinfo{author}{L.~Yu}, \bibinfo{author}{P.~Gong},
\newblock \bibinfo{title}{Google earth as a virtual globe tool for earth
  science applications at the global scale: progress and perspectives},
\newblock \bibinfo{journal}{International Journal of Remote Sensing}
  \bibinfo{volume}{33} (\bibinfo{year}{2012}) \bibinfo{pages}{3966--3986}.
\bibitem[{Chen et~al.(2015)Chen, Qiu, Wu, and Du}]{chen2015image}
\bibinfo{author}{B.~Chen}, \bibinfo{author}{F.~Qiu}, \bibinfo{author}{B.~Wu},
  \bibinfo{author}{H.~Du},
\newblock \bibinfo{title}{Image segmentation based on constrained spectral
  variance difference and edge penalty},
\newblock \bibinfo{journal}{Remote Sensing} \bibinfo{volume}{7}
  (\bibinfo{year}{2015}) \bibinfo{pages}{5980--6004}.
\bibitem[{Hu et~al.(2017)Hu, Li, Zhang, Zou, and Wu}]{hu2017unsupervised}
\bibinfo{author}{Z.~Hu}, \bibinfo{author}{Q.~Li}, \bibinfo{author}{Q.~Zhang},
  \bibinfo{author}{Q.~Zou}, \bibinfo{author}{Z.~Wu},
\newblock \bibinfo{title}{Unsupervised simplification of image hierarchies via
  evolution analysis in scale-sets framework},
\newblock \bibinfo{journal}{IEEE Transactions on Image Processing}
  \bibinfo{volume}{26} (\bibinfo{year}{2017}) \bibinfo{pages}{2394--2407}.
\bibitem[{Ronneberger et~al.(2015)Ronneberger, Fischer, and Brox}]{unet}
\bibinfo{author}{O.~Ronneberger}, \bibinfo{author}{P.~Fischer},
  \bibinfo{author}{T.~Brox},
\newblock \bibinfo{title}{U-net: Convolutional networks for biomedical image
  segmentation},
\newblock in: \bibinfo{booktitle}{Medical Image Computing and Computer-Assisted
  Intervention--MICCAI 2015: 18th International Conference, Munich, Germany,
  October 5-9, 2015, Proceedings, Part III 18},
  \bibinfo{organization}{Springer}, \bibinfo{year}{2015}, pp.
  \bibinfo{pages}{234--241}.
\bibitem[{Zhou et~al.(2018)Zhou, Rahman~Siddiquee, Tajbakhsh, and
  Liang}]{unet++}
\bibinfo{author}{Z.~Zhou}, \bibinfo{author}{M.~M. Rahman~Siddiquee},
  \bibinfo{author}{N.~Tajbakhsh}, \bibinfo{author}{J.~Liang},
\newblock \bibinfo{title}{Unet++: A nested u-net architecture for medical image
  segmentation},
\newblock in: \bibinfo{booktitle}{Deep Learning in Medical Image Analysis and
  Multimodal Learning for Clinical Decision Support: 4th International
  Workshop, DLMIA 2018, and 8th International Workshop, ML-CDS 2018, Held in
  Conjunction with MICCAI 2018, Granada, Spain, September 20, 2018, Proceedings
  4}, \bibinfo{organization}{Springer}, \bibinfo{year}{2018}, pp.
  \bibinfo{pages}{3--11}.
\bibitem[{Qin et~al.(2020)Qin, Zhang, Huang, Dehghan, Zaiane, and
  Jagersand}]{u2net}
\bibinfo{author}{X.~Qin}, \bibinfo{author}{Z.~Zhang},
  \bibinfo{author}{C.~Huang}, \bibinfo{author}{M.~Dehghan},
  \bibinfo{author}{O.~R. Zaiane}, \bibinfo{author}{M.~Jagersand},
\newblock \bibinfo{title}{U2-net: Going deeper with nested u-structure for
  salient object detection},
\newblock \bibinfo{journal}{Pattern recognition} \bibinfo{volume}{106}
  (\bibinfo{year}{2020}) \bibinfo{pages}{107404}.
\bibitem[{Wang et~al.(2022)Wang, Li, Zhang, Fang, Duan, Meng, and
  Atkinson}]{unetformer}
\bibinfo{author}{L.~Wang}, \bibinfo{author}{R.~Li}, \bibinfo{author}{C.~Zhang},
  \bibinfo{author}{S.~Fang}, \bibinfo{author}{C.~Duan},
  \bibinfo{author}{X.~Meng}, \bibinfo{author}{P.~M. Atkinson},
\newblock \bibinfo{title}{Unetformer: A unet-like transformer for efficient
  semantic segmentation of remote sensing urban scene imagery},
\newblock \bibinfo{journal}{ISPRS Journal of Photogrammetry and Remote Sensing}
  \bibinfo{volume}{190} (\bibinfo{year}{2022}) \bibinfo{pages}{196--214}.
\bibitem[{Long et~al.(2015)Long, Shelhamer, and Darrell}]{fcn}
\bibinfo{author}{J.~Long}, \bibinfo{author}{E.~Shelhamer},
  \bibinfo{author}{T.~Darrell},
\newblock \bibinfo{title}{Fully convolutional networks for semantic
  segmentation},
\newblock in: \bibinfo{booktitle}{Proceedings of the IEEE conference on
  computer vision and pattern recognition}, \bibinfo{year}{2015}, pp.
  \bibinfo{pages}{3431--3440}.
\bibitem[{Badrinarayanan et~al.(2017)Badrinarayanan, Kendall, and
  Cipolla}]{segnet}
\bibinfo{author}{V.~Badrinarayanan}, \bibinfo{author}{A.~Kendall},
  \bibinfo{author}{R.~Cipolla},
\newblock \bibinfo{title}{Segnet: A deep convolutional encoder-decoder
  architecture for image segmentation},
\newblock \bibinfo{journal}{IEEE transactions on pattern analysis and machine
  intelligence} \bibinfo{volume}{39} (\bibinfo{year}{2017})
  \bibinfo{pages}{2481--2495}.
\bibitem[{Chen et~al.(2014)Chen, Papandreou, Kokkinos, Murphy, and
  Yuille}]{deeplabv1}
\bibinfo{author}{L.-C. Chen}, \bibinfo{author}{G.~Papandreou},
  \bibinfo{author}{I.~Kokkinos}, \bibinfo{author}{K.~Murphy},
  \bibinfo{author}{A.~L. Yuille},
\newblock \bibinfo{title}{Semantic image segmentation with deep convolutional
  nets and fully connected crfs},
\newblock \bibinfo{journal}{arXiv preprint arXiv:1412.7062}
  (\bibinfo{year}{2014}).
\bibitem[{Zhao et~al.(2017)Zhao, Shi, Qi, Wang, and Jia}]{pspnet}
\bibinfo{author}{H.~Zhao}, \bibinfo{author}{J.~Shi}, \bibinfo{author}{X.~Qi},
  \bibinfo{author}{X.~Wang}, \bibinfo{author}{J.~Jia},
\newblock \bibinfo{title}{Pyramid scene parsing network},
\newblock in: \bibinfo{booktitle}{Proceedings of the IEEE conference on
  computer vision and pattern recognition}, \bibinfo{year}{2017}, pp.
  \bibinfo{pages}{2881--2890}.
\bibitem[{Li et~al.(2021{\natexlab{a}})Li, Zheng, Zhang, Duan, Wang, and
  Atkinson}]{abcnet}
\bibinfo{author}{R.~Li}, \bibinfo{author}{S.~Zheng},
  \bibinfo{author}{C.~Zhang}, \bibinfo{author}{C.~Duan},
  \bibinfo{author}{L.~Wang}, \bibinfo{author}{P.~M. Atkinson},
\newblock \bibinfo{title}{Abcnet: Attentive bilateral contextual network for
  efficient semantic segmentation of fine-resolution remotely sensed imagery},
\newblock \bibinfo{journal}{ISPRS journal of photogrammetry and remote sensing}
  \bibinfo{volume}{181} (\bibinfo{year}{2021}{\natexlab{a}})
  \bibinfo{pages}{84--98}.
\bibitem[{Li et~al.(2021{\natexlab{b}})Li, Zheng, Duan, Su, and
  Zhang}]{mresunet}
\bibinfo{author}{R.~Li}, \bibinfo{author}{S.~Zheng}, \bibinfo{author}{C.~Duan},
  \bibinfo{author}{J.~Su}, \bibinfo{author}{C.~Zhang},
\newblock \bibinfo{title}{Multistage attention resu-net for semantic
  segmentation of fine-resolution remote sensing images},
\newblock \bibinfo{journal}{IEEE Geoscience and Remote Sensing Letters}
  \bibinfo{volume}{19} (\bibinfo{year}{2021}{\natexlab{b}})
  \bibinfo{pages}{1--5}.
\bibitem[{Guo et~al.(2022)Guo, Liu, Mu, and Hu}]{eanet}
\bibinfo{author}{M.-H. Guo}, \bibinfo{author}{Z.-N. Liu},
  \bibinfo{author}{T.-J. Mu}, \bibinfo{author}{S.-M. Hu},
\newblock \bibinfo{title}{Beyond self-attention: External attention using two
  linear layers for visual tasks},
\newblock \bibinfo{journal}{IEEE Transactions on Pattern Analysis and Machine
  Intelligence} \bibinfo{volume}{45} (\bibinfo{year}{2022})
  \bibinfo{pages}{5436--5447}.
\bibitem[{Huang et~al.(2019)Huang, Wang, Huang, Huang, Wei, and Liu}]{ccent}
\bibinfo{author}{Z.~Huang}, \bibinfo{author}{X.~Wang},
  \bibinfo{author}{L.~Huang}, \bibinfo{author}{C.~Huang},
  \bibinfo{author}{Y.~Wei}, \bibinfo{author}{W.~Liu},
\newblock \bibinfo{title}{Ccnet: Criss-cross attention for semantic
  segmentation},
\newblock in: \bibinfo{booktitle}{Proceedings of the IEEE/CVF international
  conference on computer vision}, \bibinfo{year}{2019}, pp.
  \bibinfo{pages}{603--612}.
\bibitem[{Xie et~al.(2021)Xie, Wang, Yu, Anandkumar, Alvarez, and
  Luo}]{segformer}
\bibinfo{author}{E.~Xie}, \bibinfo{author}{W.~Wang}, \bibinfo{author}{Z.~Yu},
  \bibinfo{author}{A.~Anandkumar}, \bibinfo{author}{J.~M. Alvarez},
  \bibinfo{author}{P.~Luo},
\newblock \bibinfo{title}{Segformer: Simple and efficient design for semantic
  segmentation with transformers},
\newblock \bibinfo{journal}{Advances in Neural Information Processing Systems}
  \bibinfo{volume}{34} (\bibinfo{year}{2021}) \bibinfo{pages}{12077--12090}.
\bibitem[{Yang et~al.(2018)Yang, Yu, Zhang, Li, and Yang}]{denseaspp}
\bibinfo{author}{M.~Yang}, \bibinfo{author}{K.~Yu}, \bibinfo{author}{C.~Zhang},
  \bibinfo{author}{Z.~Li}, \bibinfo{author}{K.~Yang},
\newblock \bibinfo{title}{Denseaspp for semantic segmentation in street
  scenes},
\newblock in: \bibinfo{booktitle}{Proceedings of the IEEE conference on
  computer vision and pattern recognition}, \bibinfo{year}{2018}, pp.
  \bibinfo{pages}{3684--3692}.
\bibitem[{Paszke et~al.(2016)Paszke, Chaurasia, Kim, and Culurciello}]{enet}
\bibinfo{author}{A.~Paszke}, \bibinfo{author}{A.~Chaurasia},
  \bibinfo{author}{S.~Kim}, \bibinfo{author}{E.~Culurciello},
\newblock \bibinfo{title}{Enet: A deep neural network architecture for
  real-time semantic segmentation},
\newblock \bibinfo{journal}{arXiv preprint arXiv:1606.02147}
  (\bibinfo{year}{2016}).

\end{thebibliography}
	
	
	
	
	
	

\end{document}